\documentclass{article}

\PassOptionsToPackage{numbers,sort&compress}{natbib}
\usepackage[preprint]{neurips_2025}  %

\input{preamble}

\title{
    Back to the Features: DINO as a \\
    Foundation for Video World Models
}

\author{%
\bf Federico Baldassarre
\And
\bf Marc Szafraniec
\And
\bf Basile Terver
\And
\bf Vasil Khalidov
\AND
\bf Francisco Massa
\And
\bf Yann LeCun
\And
\bf Patrick Labatut
\And
\bf Maximilian Seitzer
\And
\bf Piotr Bojanowski
\AND
Meta FAIR
}

\begin{document}
\maketitle

\begin{abstract}
We present DINO-world, a powerful generalist video world model trained to predict future frames in the latent space of DINOv2.
By leveraging a pre-trained image encoder and training a future predictor on a large-scale uncurated video dataset, DINO-world learns the temporal dynamics of diverse scenes, from driving and indoor scenes to simulated environments.
We show that DINO-world outperforms previous models on a variety of video prediction benchmarks, \eg segmentation and depth forecasting, and demonstrates strong understanding of intuitive physics.
Furthermore, we show that it is possible to fine-tune the predictor on observation-action trajectories.
The resulting action-conditioned world model can be used for planning by simulating candidate trajectories in latent space.
\end{abstract}

\section{Introduction}
\label{sec:intro}

In 2018, \citet{ha2018world} popularized the concept of a \emph{world model}, a neural network that predicts the future state of an environment given past observations and actions taken by an agent.
Recently, the subject of world models has gained traction~\citep{hu2023gaia,yang2023learning,brooks2024video,bruce2024genie,parkerholder2024genie2,bartoccioni2025vavam,agarwal2025cosmos,russell2025gaia2}, with conditional generative models showing impressive results on specialized domains such as driving~\citep{hu2023gaia,bartoccioni2025vavam,russell2025gaia2}, or video games~\citep{bruce2024genie,parkerholder2024genie2}.
Likewise, large-scale generative video models with other kinds of conditioning, such as SORA~\citep{brooks2024video}, MovieGen~\citep{polyak2025moviegen}, or Wan2.1~\citep{wang2025wan}, can be considered world models.
Even \emph{unconditional} generative models can be \emph{post-trained} with action data to enable control, as proposed \eg in COSMOS~\citep{agarwal2025cosmos}.
The common thread that ties together these works is the pre-training of video world models on large-scale datasets, which emerges as the next frontier of unsupervised learning.

Training effective world models holds significant promise for advancing artificial intelligence. 
The hypothesis is that by predicting the evolution of the physical world in all its complexity,
these models acquire a deep understanding of reality~\citep{lecun2022path}.
Common sense knowledge grounded in reality may help fill the gaps that large language models still exhibit~\citep{majumdar2024OpenEQA,tong2024eyeswideshut}.
More concretely, world models can be used to control agents, both by training policies with simulated experience~\citep{Hafner2023MasteringDD,alonso2025diffusion} and by planning online~\citep{Hansen2023TDMPC2,zhou2024dinowm,sobal2025learning}.
This increases sample efficiency and reduces the amount of potentially expensive or risky interactions with the environment.
Crucially, planning also allows for zero-shot adaptation to new tasks and environments, enabling generalization beyond the training data~\citep{sobal2025learning}.

However, training effective world models is challenging.
First, there is a data problem: large-scale high-quality video collections are costly to obtain, especially if annotated with actions, and often require proprietary data.
The action space is also task-dependent, which complicates generalization across different applications.
For this reason, current successful applications of world models are limited to restricted domains such as self-driving or games~\citep{hu2023gaia,alonso2025diffusion}.
Second, the task itself is hard: accurately modeling physics and behaviors in unconstrained, partially observable environments is an unsolved problem, even over short time horizons. %
So far, the best pixel-based generative models are enormously resource hungry, \eg 22M GPU-hours for COSMOS~\citep{agarwal2025cosmos}.
Yet, this capacity might be wasted on irrelevant details; for example, predicting the exact motion of each leaf in the wind for a system designed for autonomous driving.
Capturing the environment at an \emph{appropriate level of detail} becomes therefore crucial for the effectiveness and efficiency of the model.
However, the unsupervised learning of general abstractions that facilitate both future prediction and controlling agents remains an open problem~\citep{lecun2022path}.
Last, the evaluation of pre-trained video world models is a challenge on its own.
Previous works on world models have focused on a narrow set of benchmarks depending on the application---generation quality~\citep{yang2023learning,wang2025wan}, forecasting accuracy~\citep{karypidis2024dinoforesight}, action physical understanding~\citep{garrido2025intuitive,motamed2025generative}, or agent control~\citep{zhou2024dinowm}---but more holistic evaluations are missing.  %

\looseness-1
In this work, we take a step towards addressing these challenges.
In particular, we propose to \emph{pre-train a video world model in the latent space of a frozen vision encoder} such as DINOv2~\citep{oquab2024dinov2}. 
The model can then be \emph{post-trained} on actions for planning and control.
This approach has several advantages: 
1) the separation of video \emph{pre-training} from action-conditioned fine-tuning enables learning general knowledge from large pools of unlabeled videos, later reducing the demand for data labeled with actions,
2) training a \emph{latent} world model sidesteps the challenges of modeling individual pixels, unnecessary for most downstream tasks,
3) the \emph{frozen encoder} bootstraps the learning process by reusing the strong semantic and geometric understanding of DINO, and avoids the technical complexity of training encoder \& predictor jointly~\citep{bardes2024revisiting}.
Furthermore, we introduce a streamlined world model architecture that is vastly more resource efficient than state-of-the-art models~\citep{agarwal2025cosmos}, both in training and inference. %
We train the predictor on a large-scale uncurated dataset of \({\sim}60\)M web videos, allowing it to acquire general features that transfer well across diverse domains.  %
In our experimental evaluations, we strive to cover a wide range of tasks---from dense forecasting to intuitive physics---and to bring together the many kinds of world models that have been proposed in the literature.
In VSPW segmentation forecasting, our model achieves $6.3\%$ higher mIoU than the second best model when predicting $0.5$ seconds into the future.  %
When post-training on action data and evaluating on planning tasks, our experiments confirm the advantage of large-scale unsupervised pre-training.

Our contributions can be summarized as follows:
\textbf{(i)}
We propose a simple architecture for training a world model by predicting future visual features.
Our model handles variable frame rates, context lengths, and resolutions.
\textbf{(ii)}
We train our predictor at scale, with model sizes up to ViT-g, on a dataset of more than 60M videos.
\textbf{(iii)}
We design a comprehensive evaluation for our method, covering a wide range of tasks, and comparing with several world models from different paradigms.
\textbf{(iv)}
We show on several planning tasks, that our unconditional world model can be easily fine-tuned with actions, demonstrating an advantage over training from scratch.

\vspace{-0.5em}
\section{Related work}
\label{sec:related}
\vspace{-0.5em}

World models are an active area of research.
This section attempts to clarify definitions and organize the heterogeneous landscape of world models.
At a high level, we define world models as models capable of predicting the temporal evolution of an environment given past visual observations and an optional conditioning signal.
The predictions can be in observation space, \eg pixels, or in a latent space; the additional information for conditioning can include agent actions, language instructions, or other constraints.
In \Cref{subsec:world-models-applications}, we give an overview of the application areas of world models; in \Cref{subsec:world-models-design-choices}, we discuss several design choices.

\vspace{-0.5em}
\subsection{Applications}
\label{subsec:world-models-applications}
\vspace{-0.5em}

\myparagraph{Controlling agents.} 
Originally, world models were introduced to enable agents to understand and achieve goals within their environment. 
Prior to Ha's and Schmidhuber's~\citep{ha2018world} eponymous work, learned models have long been used in model-based reinforcement learning (RL) and adaptive control~\citep{Sutton1991Dyna,Schmidhuber1990MakingTheWorld,goodwin1984adaptive}.
The two primary approaches of using a learned model to control an agent are
via \emph{training in simulation}~\citep{ha2018world,Hafner2019DreamTC,Hafner2023MasteringDD,micheli2023transformers,alonso2025diffusion}, where a policy is trained on ``imagined'' trajectories using model-free RL without environment interactions,
or via goal-based \emph{planning}~\citep{Hafner2019plannet,Hansen2023TDMPC2,bar2024navigationworldmodels,zhou2024dinowm}, where the model simulates candidate trajectories at inference time to identify a viable sequence of actions.  %

\myparagraph{Generating videos.}
More recently, the term world model has broadened to include (controllable) generative models of video trained on large-scale data~\citep{hu2023gaia,yang2023learning,brooks2024video,bruce2024genie,parkerholder2024genie2,bartoccioni2025vavam,agarwal2025cosmos}. %
This line of research continues a rich literature on video generation with neural networks, from unconditional~\citep{kalchbrenner2016videopixelnetworks,babaeizadeh2018stochastic,denton2018stochasticvideogen,yan2021videogpt,yu2023magvit}, to language-~\citep{Gupta2023PhotorealisticVG,villegas2023phenaki,Kondratyuk2024VideoPoetAL} and action-conditional~\citep{oh2015actioncondvideopred,chiappa2017recurrentenvsims}.  %
Conditioned on textual prompts and/or visual input, generative world models are able to produce plausible videos, often allowing fine-grained control, \eg directing the camera~\citep{agarwal2025cosmos} or a simulated actor~\citep{bruce2024genie,parkerholder2024genie2,agarwal2025cosmos}.  %
Although recent models can generate high-fidelity simulations, most remain limited to their training domains, \eg driving or games~\citep{hu2023gaia,bartoccioni2025vavam,russell2025gaia2,bruce2024genie,parkerholder2024genie2}.
Furthermore, their physical understanding appears to be limited currently~\citep{Kang2024HowFar,motamed2025generative}.
While there have been early demonstrations of training robot policies in simulated videos~\citep{yang2023learning}, the utility of generative models for controlling agents is generally unclear.

\myparagraph{Forecasting.}
Predicting modalities other than pixels may also be of interest, \eg semantic~\citep{luc2017predsemantic,chiu2019segmentingfuture,karypidis2024dinoforesight} or instance segmentation~\citep{luc2018predinstance,karypidis2024dinoforesight}, depth maps~\citep{karypidis2024dinoforesight}, or even semantic attributes such as human actions~\citep{vondrick2016anticipating,girdhar2021anticipativevideotransformer,zhong2022anticipativefeaturefusiontransformer}.
Structurally, these approaches resemble unconditional world models in that they roll out latent features into the future, then apply decoders to obtain the property of interest.
Taking inspiration from these approaches, we consider tasks such as semantic segmentation and depth forecasting as a means to probe the quality of our proposed world model.

\vspace{-0.5em}
\subsection{Design choices}
\label{subsec:world-models-design-choices}
\vspace{-0.5em}

\myparagraph{Encoder and latent space.}
Nearly all world models are based on encoding image frames into a compressed latent space, where the future predictions take place.
The choice of latent space, that is, the \emph{encoder} or \emph{tokenizer}, introduces various trade-offs and depends on the intended purpose of the world model.
Video generation models typically opt for continuous or discrete VAEs~\citep{kingma2022vae,oord2018vqvae} to achieve high visual fidelity. 
In contrast, world models intended for downstream tasks such as controlling agents or predicting future segmentation maps benefit from predicting the future in a high-level, semantic latent space~\citep{lecun2022path,zhou2024dinowm,karypidis2024dinoforesight}. %
For this purpose, vision foundation models~\citep{oquab2024dinov2,tschannen2025siglip2multilingualvisionlanguage} that already acquired a broad knowledge from image pretraining are prime candidates.  %
In this paper, we similarly argue for learning \emph{latent-space world models} on top of pre-trained image models.

\looseness-1
\myparagraph{Multi-stage training.}
We compare two strategies for building world models:
1) first training an unconditional model, and then finetuning it with action- or instruction-conditioning~\citep{agarwal2025cosmos,bartoccioni2025vavam}, or
2) directly training a conditional world model~\citep{hu2023gaia,Hafner2023MasteringDD,micheli2023transformers,zhou2024dinowm,alonso2025diffusion}.
The latter is advantageous when sufficient annotated data is available.
However, for many problems, data is sparse, and pre-training lets the model acquire general knowledge about the world.
In this work, we focus on this \emph{unconditional pretraining} stage and show that high-quality world models can be learned from large video collections.
A related question is whether to train the visual encoder jointly with the world model or in an separate stage.
Joint training is used successfully in narrow environments~\citep{Hafner2019plannet,Hafner2019DreamTC,Hafner2023MasteringDD,Hansen2023TDMPC2}, while separate training is advantageous when tackling broad complex domains thanks to the separation of concerns between visual and dynamics modeling~\citep{agarwal2025cosmos}.
Notably, V-JEPA jointly trains an encoder and a predictor at scale~\citep{bardes2024revisiting}, obtaining features suitable for video summarization, but suboptimal for forecasting and planning, as shown in \Cref{tab:dense-forecasting}.
In this work, we choose to decouple visual representation learning from world modeling and benefit from pre-trained foundation models.

\myparagraph{Frameworks.}
Several modeling approaches have been proposed for world modeling, ranging from recurrent state-space models~\citep{Hafner2023MasteringDD,Hansen2023TDMPC2}, to diffusion~\citep{yang2023learning,brooks2024video,agarwal2025cosmos}, auto-regressive models~\citep{hu2023gaia,Kondratyuk2024VideoPoetAL,agarwal2025cosmos,bartoccioni2025vavam}, and masked reconstruction models~\citep{yu2023magvit,bardes2024revisiting}.
We refer to the work of \citet{zhu2024worldmodelsurvey} for an overview. 
In this work, we introduce a transformer-based auto-regressive architecture that is not bound to fixed prediction intervals and allows us to flexibly trade-off prediction quality and inference cost.

\vspace{-0.5em}
\section{Method}
\label{sec:method}
\vspace{-0.5em}

\begin{figure}[t]
    \centering
    \includegraphics[width=1\linewidth]{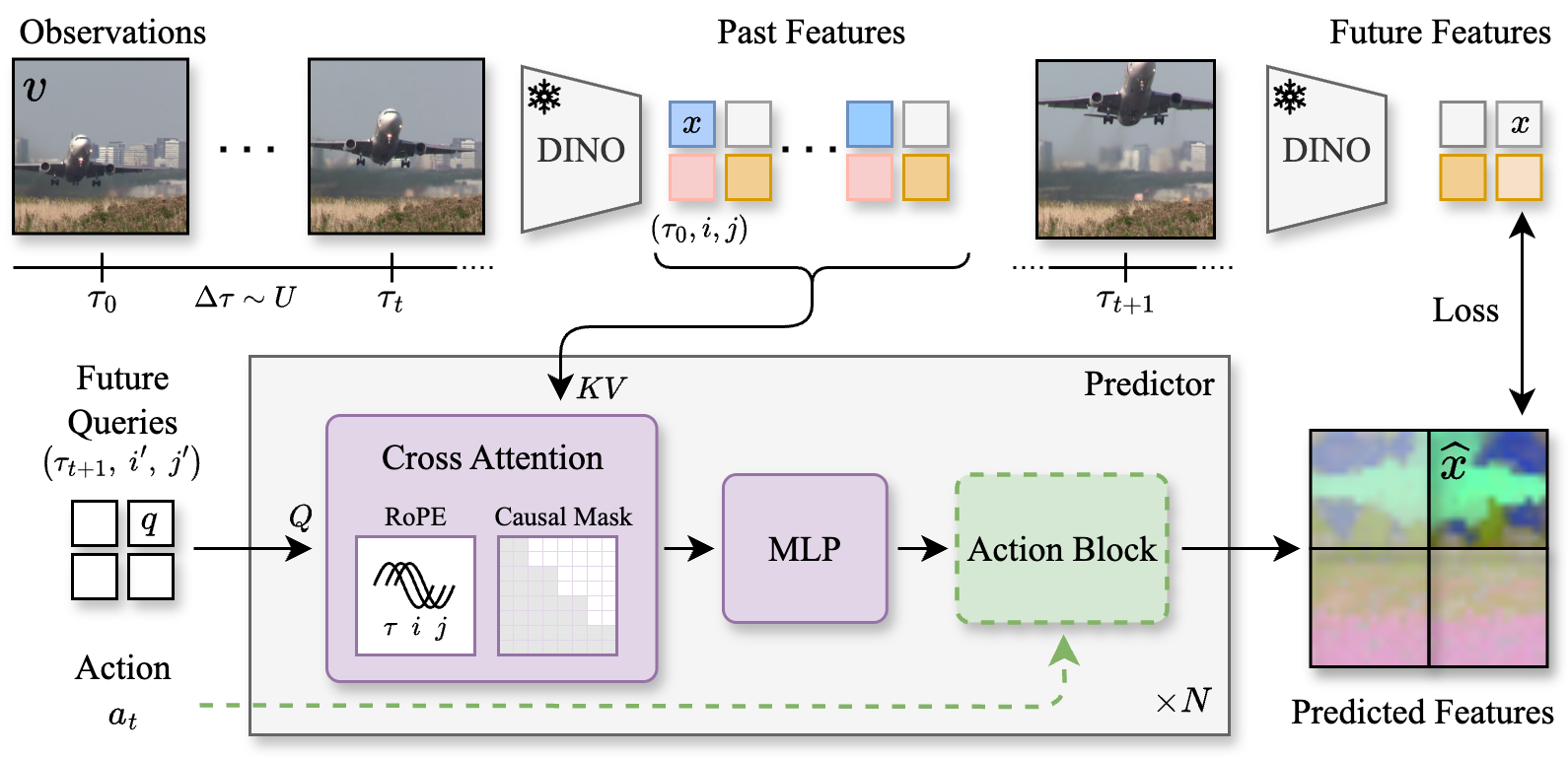}
    \caption{
        \textbf{Latent video world model architecture.}
        A frozen DINOv2 \emph{encoder} maps video frames to patch tokens in latent space.
        The \emph{predictor} is a stack of cross-attention blocks trained to predict the patch token \(\hat{\vx}_{t+1,i',j'}\) at a future timestamp \(\tau_{t+1}\) and location \((i',j')\) from all past tokens \(\vx_{1:t,\cdot,\cdot}\) and timestamps \(\bm{\Tau}_{1:t}\).
        Spatiotemporal coordinates are injected into each token via a 3-axial RoPE.
        The block-triangular attention mask allows to train the model in parallel for all patches of all frames, respecting temporal causality.
        For conditional fine-tuning, we add zero-initialized action blocks that update each query with the corresponding action \(\va_t\).
    }
    \label{fig:architecture-diagram}
    \vspace{-0.5em}
\end{figure}

We wish to train a world model capable of understanding the temporal dynamics of real-world videos.
\Cref{fig:architecture-diagram} outlines the main components of our method, a \emph{frame encoder} and a \emph{future predictor}.
In \Cref{sec:method-input-state-representation}, we introduce notation for the observation space, \ie video pixels, and we identify the state representation to be modeled, namely patch features in the latent space of DINOv2.
Then, in \cref{sec:method-world-model-architecture}, we describe the architecture and training procedure of our world model, \ie a cross-attention decoder that learns to predict future states from past observations.
Finally, \Cref{sec:method-actions} describes how to fine-tune the model with available conditioning data, \eg trajectories with agent actions.

\vspace{-0.5em}
\subsection{Observations and state space}
\label{sec:method-input-state-representation}
\vspace{-0.5em}

\myparagraph{Observations.}
Digital videos, captured at a certain resolution and frame rate, consist of a sequence of RGB frames and their associated timestamps.
We represent a video of \(T\) frames at \(H'{\times}W'\) resolution as a sequence of frames and timestamps
\(\Set{(\vv_t, \tau_t)}_{t=1}^T\) with \(\vv_t \in \Real^{H'{\times}W'{\times}3}\) and \(\tau_t \in \Real_+\),
or, equivalently, as the tensors \(\mV \in \Real^{T{\times}H'{\times}W'{\times}3}\) and \(\bm{\Tau} \in \Real_+^{T}\).
A common simplification in the literature \citep{bardes2024revisiting,karypidis2024dinoforesight} is to assume a fixed frame rate (FPS), omit the timestamps \(\tau\), and model the discrete sequence indexed by \(t\).
In this work, we explicitly model time to accommodate videos at variable FPS during training and to unlock fine-grained temporal control for inference.

\myparagraph{Frame encoder.}
Instead of directly modeling pixels, virtually all modern world models operate on a latent representations of video patches.
V-JEPA contains an \emph{encoder} and a \emph{predictor}, optimized jointly  \citep{bardes2024revisiting}.
Even generative models, such as SORA \citep{brooks2024video} and COSMOS \citep{agarwal2025cosmos}, train their predictors in the latent space of a pre-trained autoencoder, whose tokens are eventually decoded to pixels.
Such models optimize for generation quality, and therefore leverage autoencoders with low reconstruction error.
In contrast, we choose to encode video frames with a foundation model designed for representation learning and trained with self supervision, namely DINOv2 \citep{oquab2024dinov2}.
We posit that working in such a latent space will drastically reduce the compute required to train the predictor, having already learned strong semantic features of still images.
Indeed, we demonstrate effective training of world models with \({<}1\) billion parameters, while state-of-the-art generative models like COSMOS can use up to 12 billions \citep{agarwal2025cosmos}.
While the predictions of our world model can not be rendered directly as pixels, they can readily be used in dense prediction tasks (see \Cref{sec:experiments-intphys-and-forecasting}).

\myparagraph{State.}
DINOv2 maps each frame to a feature tensor
\(\vx_t = \Encoder{\vv_t} \in \Real^{H{\times}W{\times}D}\),
where \(D\) is the embedding dimension, and \(H{\times}W\) the resulting spatial resolution.
For each timestamp \(\tau_t\) and spatial location \((i, j)\), the \emph{patch token} \(\vx_{t,i,j}\) represents the unit of \emph{state} that the world model will learn to predict.
With visual observations embedded as vectors in the latent space of DINOv2, we can formally define a world model as the mapping:
\begin{equation}
\label{eq:world-model-predictor}
    \left(\mathbf{X}_{1:t}, \bm{\Tau}_{1:t}, (\tau_{t'}, i', j') \right) \rightarrow \vx_{t',i',j'} \qquad
    \forall (i',j') \in \{1, \ldots, H\}{\times}\{1, \ldots, W\},\ \forall t' > t,
\end{equation}
where \(\mX_{1:t}\) and \(\bm{\Tau}_{1:t}\) are past features and timestamps, and \((\tau_{t'}, i', j')\) are the future coordinates for which a patch token should be predicted.
In the next section, we will discuss the architecture, the loss function, and the optimization procedure to train such a predictor.

\vspace{-0.4em}
\subsection{Predictor architecture and training}
\label{sec:method-world-model-architecture}
\vspace{-0.4em}

\myparagraph{Architecture.}
Taking inspiration from neural machine translation \citep{vaswani2017attention} and image reconstruction \citep{fu2024rethinking}, we frame the prediction task as a decoding problem, and design the predictor as a stack of \(N\) residual pre-norm cross-attention blocks.
To predict the future state at coordinates \((\tau_{t'},i',j')\), we initialize a \emph{query} token \(\vq\in\Real^{D'}\) from a learnable embedding.
At each block, the query token cross-attends to \emph{key-value} pairs obtained from all previous patch tokens, followed by an MLP block:
\begin{align}
    \vq &\leftarrow \vq + \CrossAttention{\LayerNorm{\vq}}{\SetDef{\vx_{t,i,j}}{\tau_t<\tau_{t'}}}, \label{eq:cross-attn-block}\\
    \vq &\leftarrow \vq + \Mlp{\LayerNorm{\vq}}. \label{eq:mlp-block}
\end{align}
After the last block, a linear projection maps \(\vq\) to obtain the predicted patch token \(\hat{\vx}_{t',i',j'}\in\Real^D\).

\myparagraph{Positional encoding.}
With this formulation, the query vector \(\vq\) and the context features \(\vx\) do not carry information about their location in the video.
To allow the model to reason about the spatial and temporal relationships between tokens, we inject rotary position encodings (RoPE) \citep{su2024roformer} into the multi-head attention.
Specifically, we split the head dimension \(D_h\) in three, and encode the temporal, horizontal, and vertical coordinates of each token separately.
For the spatial coordinates \((i, j)\), we use relative positions defined on a \([-1, +1]^2\) grid, such that varying the input resolution will not change the relative distance between patches.
For the temporal coordinate \(\tau\), we use absolute timestamps in seconds, so that the model can distinguish between high and low frame rates, and extrapolate to longer videos.
Considering the input space, we use RoPE periods in the range \([10^{-2}, 10^2]\).

\myparagraph{Training objective.}
In principle, the world model in \cref{eq:world-model-predictor} could predict at any timestamp \(\tau_{t'}\) in the future.
For ease of parallelization, we train the model with a next-frame prediction objective, \ie \(t' = t + 1\), and teacher forcing.
Given a sequence of \(T\) frames, we compute all predictions \(\hat{\vx}_{t+1, i', j'}\) for all \((i', j')\) and \(t \in {\Set{1, \ldots, T{-}1}}\) in parallel by stacking \((T{-}1)HW\) queries and masking the attention with a block-triangular pattern to preserve causality, \ie a query for frame \(t+1\) can only attend to patch tokens up to frame \(t\).
For a predictor parameterized by \(\theta\), the training objective is:
\begin{equation}
\label{eq:training-objective}
    \min_\theta \Loss\bigl(
        \vx_{t+1,i',j'},
        \textsc{Predictor}_\theta\bigl(
            \mX_{1:t}, \bm{\Tau}_{1:t}, (\tau_{t+1}, i', j')
        \bigl)
    \bigl) \qquad
    \forall i',j',t \in \Set{1, \ldots, T{-}1},
\end{equation}
where \(\Loss\) is a smooth L1 loss and the optimization is performed over all \((T-1)HW\) terms in parallel.
In contrast, masked reconstruction losses, \eg V-JEPA or DINO-Foresight, only compute a loss term for the \emph{mask tokens}, which are a small fraction of all processed tokens.

\myparagraph{Variable FPS.}
During optimization, a naive strategy for sampling training clips from a video is to take \(T\) contiguous frames with a random starting point.
This skews the distribution of time deltas \(\Delta\tau = \tau_{t+1} - \tau_t\) towards short intervals, which in turn limits the predictive horizon of a model trained with \cref{eq:training-objective}.
Instead, we would like the model to predict at any \(\Delta\tau\) in the future, within reasonable limits.
Therefore, for every video, we sample \(T{-}1\) time deltas uniformly from a predefined range \([\Delta\tau_\text{min}, \Delta\tau_\text{max}]\), and compute \(T\) timestamps by taking their cumulative sum and a random starting point.
For each sampled timestamp, we then decode the nearest frame and its actual timestamp, used for training.
This ensures that the model is trained on a uniform distribution of time intervals.

\subsection{Action-conditioned fine-tuning}
\label{sec:method-actions}

Our video world model can be trained with self-supervision on large-scale collections of unlabeled videos.
Still, many downstream applications involve a conditioning signal, \eg agent actions or language instructions, for which data is likely limited.
In this work, we focus on agent trajectories in the form of observation-action pairs \((\vv_t, \va_t)\), which is a common setting in robotics, autonomous driving, and reinforcement learning.
Starting from a pretrained video world model, we propose a simple adaptation to condition the prediction for frame \(t{+}1\) on the action \(\va_t\).
After each block (\cref{eq:mlp-block}), we add an \emph{action block} that updates the query as \(\vq + \Mlp{\LayerNorm{[\vq, \va]}}\) using the corresponding action.
These action blocks can be initialized as an identity and trained with a small dataset of action-conditioned trajectories.
Optionally, the video world model can remain frozen, training only the action blocks, which alleviates overfitting and allows reusing the same base model for different tasks.
The alternative approach of interleaving action tokens to the sequence of patch tokens, as done in \eg DINO-WM \citep{zhou2024dinowm}, has several drawbacks.
Mixing tokens of different types complicates batching and masking, requires additional capacity in the model, and forces a full fine-tuning of the whole model, potentially destroying the learned video understanding.

\section{Experiments}
\label{sec:experiments}

In this section, we empirically validate the quality of our world model.
In \Cref{sec:experiments-intphys-and-forecasting}, we evaluate the latent predictions of an unconditional model on dense feature forecasting tasks and on three physics understanding benchmark.
Then, in \Cref{sec:experiments-ablations}, we analyze the effect of the different components of our model.
Finally, \Cref{sec:experiments-action-conditioning} demonstrates fine-tuning the world model on agent trajectories and applying it to planning on three simulated RL environments.

\myparagraph{Implementation.}
We encode frames using a DINOv2 ViT-B/14 with registers \citep{oquab2024dinov2,darcet2023registers},
the same encoder used in DINO-Foresight \citep{karypidis2024dinoforesight}, though we only model the features of the last layer (\(D{=}768\)).
The predictor is a transformer of \(N{=}40\) blocks, dimension \(D'{=}1536\), and \(24\) heads, akin to a ViT-g, but with cross attention.
We train our model with AdamW~\citep{loshchilov2019decoupled} for 300k iterations with batches of 1024 clips, \(T{=}8\) and resolution \(224{\times}224\); followed by 50k iterations at \(448{\times}448\).
After a short warmup, the learning rate remains constant at \(10^{-4}\).
All hyperparameters are reported in \Cref{appendix:world-model-details}.

\myparagraph{Training data.}
We use a large-scale private pool of \({\sim}66\)M uncurated videos with duration of 5--60 seconds and different frame rates.
As part of our ablations (see \cref{sec:experiments-ablations}), we also train on open-access video datasets such as Cityscapes~\citep{cordts2016cityscapes} and Something-Something V2 \citep{goyal2017something}, whose smaller size and narrow domain yield lower performance.
We provide additional dataset details in \Cref{appendix:datasets}. 

\myparagraph{Baselines.}
We compare our method with recent \sota world models, both pixel- and latent-space based.
\textbf{DINO-Foresight}~\citep{karypidis2024dinoforesight} is a predictive model similar to ours, but trained with a masked reconstruction objective on a narrow domain, \ie Cityscapes~\citep{cordts2016cityscapes}.
\textbf{V-JEPA}~\citep{bardes2024revisiting} is a self-supervised video representation model that jointly optimizes the encoder and predictor through a masked reconstruction objective.
Note that V-JEPA's predictor is a ``head'' used to train the encoder, and was not designed to provide accurate future predictions.
V-JEPA is trained on VideoMix2M, a mix of three datasets: HowTo100M~\citep{Miech2019HowTo100MLA}, Kinetics 400/600/700~\citep{kay2017kinetics}, and SSv2~\citep{goyal2017something}.
\textbf{COSMOS}~\citep{agarwal2025cosmos} is a pixel-space world model capable of generating videos conditioned on past frames.
COSMOS models are significatively larger, 4--12B parameters, and are trained on a private dataset of 100M carefully curated videos.
Despite the common denomination of ``world models'', these baselines stem from different paradigms and goals.
Therefore, we highlight the importance of bringing them together under a comprehensive evaluation.

\subsection{Dense feature forecasting and intuitive physics}
\label{sec:experiments-intphys-and-forecasting}

\begin{table*}[b]
    \centering
    \vspace{-1em}
    \caption{
        \textbf{Dense forecasting.} 
        For each method, we train a ``present-time'' linear head for segmentation or depth estimation, which we then apply to the output of the world model at a prediction distance of \({\sim}200\)ms (short) or \({\sim}0.5\) seconds (mid).
        A small gap between present and forecasting performance indicates a strong world model.
    }
    \label{tab:dense-forecasting}
    \resizebox{1.0\textwidth}{!}{%
        \begin{tabular}{@{}ll ccc ccc ccc @{}}
            \toprule
            & & \multicolumn{3}{c}{VSPW mIoU (\(\uparrow\))} & \multicolumn{3}{c}{Cityscapes mIoU (\(\uparrow\))} & \multicolumn{3}{c}{KITTI RMSE (\(\downarrow\))} \\
            \cmidrule(lr){3-5} \cmidrule(lr){6-8} \cmidrule(lr){9-11}
            & Encoder            & Present & Short & Mid & Present & Short & Mid & Present    & Short    & Mid   \\
            \midrule
            Copy Last   & ViT-B     & \textcolor{mydarkgray}{52.8} & 47.9 & 42.1 & \textcolor{mydarkgray}{68.6} & 53.2 & 39.7 & \textcolor{mydarkgray}{2.963} & 3.778 & 4.745 \\
            \midrule                                                        
            COSMOS-4B & ViT-B         & \textcolor{mydarkgray}{52.8} & 46.6 & 40.2 & \textcolor{mydarkgray}{68.6} & 55.4 & 46.2 & \textcolor{mydarkgray}{2.963} & 4.178 & 4.742    \\
            COSMOS-12B & ViT-B        & \textcolor{mydarkgray}{52.8} & 46.6 & 40.7 & \textcolor{mydarkgray}{68.6} & 55.6 & 45.9 & \textcolor{mydarkgray}{2.963} & 4.157 & 4.617    \\
            \midrule
            V-JEPA & ViT-L          & \textcolor{mydarkgray}{29.1} & \phantom{0}8.2 & \phantom{0}7.7 & \textcolor{mydarkgray}{48.8} & 15.5 & 14.0 & \textcolor{mydarkgray}{3.502} & 7.217 & 7.491 \\
            V-JEPA & ViT-H          & \textcolor{mydarkgray}{28.0} & \phantom{0}4.9 & \phantom{0}4.6 & \textcolor{mydarkgray}{49.7} & 13.3 & 12.2 & \textcolor{mydarkgray}{3.402} & 5.458 & 5.785 \\                                                        
            DINO-Foresight & ViT-B  & \textcolor{mydarkgray}{50.6} & 44.7 & 37.7 & \textcolor{mydarkgray}{66.9} & 64.5 & \textbf{57.2} & \textcolor{mydarkgray}{2.882} & 3.562 & \textbf{3.740} \\
            \midrule                                                        
            DINO-world & ViT-B      & \textcolor{mydarkgray}{52.8} & \textbf{51.6} & \textbf{47.0} & \textcolor{mydarkgray}{68.6} & \textbf{64.7} & 55.1 & \textcolor{mydarkgray}{2.963} & \textbf{3.214} & 4.268 \\
            \bottomrule
        \end{tabular}
    }
\end{table*}

\myparagraph{Dense forecasting.}
In line with previous work \citep{karypidis2024dinoforesight,luc2017predsemantic}, we assess the quality of the predicted features using dense forecasting.
This is a proxy task that allows us to compare future predictions across models through the lens of a downstream task.
Specifically, given past frames as context, we evaluate the model's ability to predict segmentation and depth for a future timestamp. 
The segmentation and depth models are first trained on present data and subsequently applied to features generated for the near future.
We run this evaluation on Cityscapes~\citep{cordts2016cityscapes}, VSPW~\citep{miao2021vspw}, and KITTI~\citep{geiger2013vision}.
Following the setup of \citet{luc2017predsemantic}, we evaluate the model in two settings:
for \emph{short-term} forecasting, the model is requested to predict \({\sim}200\)ms in the future,
while for \emph{mid-term} forecasting, the prediction target is at \(0.5\) seconds.
For our model, we consider temporally-aligned pairs of images and labels, and train a ``present-time'' linear head for each task on top of features extracted from the DINOv2 encoder.
At inference time, given four past frames, the world model predicts features at a future timestamp, to which we directly apply the heads to obtain a prediction.
For COSMOS, we encode the generated frame using DINOv2 and apply the same linear heads used on our model. 
This setup allows for a direct comparison between the two paradigms: predicting latent features versus generating pixels.
For V-JEPA and DINO-Foresight, we train ad-hoc linear heads, to account respectively for the jointly-learned encoder and the training aspect ratio.
We also consider a simple setup where the features of the last observed frame are treated as future predictions, setting a baseline score for the three datasets.
For more details on the experimental setup, please refer to the appendix.
The results in \Cref{tab:dense-forecasting} demonstrate that our world model surpasses joint-predictor architectures like V-JEPA and generative models like COSMOS. 
The marginally superior performance of DINO-Foresight on Cityscapes and KITTI can be attributed to its domain-specific training, \ie driving videos.
The strong performance of our model across those benchmarks validates the proposed paradigm: training latent-space world model atop a frozen SSL encoder.
In fact, the predicted features are of higher quality compared to V-JEPA, and video dynamics are modeled more accurately, compared to COSMOS.

\myparagraph{Intuitive physics.}
We expect that a well-trained world model demonstrates an adequate understanding of the physical world, and can be used to estimate the plausibility of a video stream.
To assess this ability, we consider three \emph{intuitive physics} benchmarks: IntPhys \citep{riochet2021intphys}, GRASP \citep{jassim2024grasp}, and InfLevel \citep{weihs2022benchmarking}.
Following the experimental protocol proposed by~\citet{garrido2025intuitive}, we define a \emph{surprise} score based on the model's predictions.
This score quantifies the deviation from expected physical behavior.
Intuitively, the prediction error of a model should be low for plausible videos and higher for implausible ones, \eg scenes that violate physical principles such as object permanence or gravity.
This is similar in spirit to likelihood-based evaluations of LLMs.
For our world model, V-JEPA, and DINO-Foresight, we use the mean absolute error between predicted and actual features as a measure of surprise.
For COSMOS, we use the perplexity of the autoregressive predictor as it predicts discrete latent tokens.
From the results in \Cref{tab:intphys-compact-results}, we observe that all world models trained on large-scale datasets achieve an adequate level of physics understanding.
Our model achieves comparable performance to V-JEPA ViT-H, despite using a smaller encoder.
The lower performance of DINO-Foresight on IntPhys and GRASP can be attributed to its training domain, which does not contain synthetic videos.
COSMOS achieves near-perfect performance on the simpler IntPhys, while falling short on the other two.
However, these tasks are inherently noisy due to significant distribution shifts and the extensive temporal context required to accurately assess plausibility, which can be challenging.
Therefore, we treat these evaluations as a sanity check rather than a benchmark.
In the following section, we will ablate several design choices of our approach.

\begin{table*}[b]
    \vspace{-1em}
    \centering
    \caption{
        \textbf{Intuitive physics benchmarks.}
        For each dataset, we identify physically implausible videos by tracking the amount of ``surprise'', \ie the difference between encoded and predicted features.
        We report the mean relative accuracy across all video categories.
        The detailed breakdown per categories is provided in the appendix.
    }
    \label{tab:intphys-compact-results}
    {
    \small
    \begin{tabular}{@{}lrr ccc@{}}
        \toprule
                & Encoder & Predictor & IntPhys & GRASP & InfLevel \\
        \midrule
        COSMOS-4B      & VAE    & 4B & \textbf{99.5} & 60.1 & 44.8 \\  %
        \midrule
        V-JEPA         & ViT-L  & \phantom{0}22M & 92.2 & 67.0 & 58.9 \\  %
        V-JEPA         & ViT-H  & \phantom{0}22M & 89.4 & 73.0 & 59.9 \\  %
        DINO-Foresight & ViT-B  & 193M & 87.8 & 64.9 & 62.8 \\
        \midrule
        DINO-world     & ViT-B  & 1.1B & 91.3 & \textbf{76.0} & \textbf{63.7} \\ %
        \bottomrule
    \end{tabular}
    }
\end{table*}

\subsection{Ablation studies}
\label{sec:experiments-ablations}

\myparagraph{Predictor size.}
For the main results in \Cref{sec:experiments-intphys-and-forecasting}, we applied a ``giant'' predictor to a relatively smaller ViT-B encoder.
We ablate this choice by training predictors in ``base'', ``large'' and ``giant'' sizes for 300k iterations at \(224{\times}224\) resolution.
We report performance on IntPhys, mid-term Cityscapes and VSPW in \Cref{tab:ablation-results} (left).
The results demonstrate a clear scaling trend, with larger predictors yielding better performance on all tasks.
This suggests that modeling temporal dynamics requires larger capacity than modeling spatial features in static images.

\myparagraph{Training data.}
In this study, we evaluate the impact of training data, in particular the size and diversity of the dataset.
We train our default model on two additional datasets, Cityscapes and Something-Something V2 (SSv2), which cover the narrow domains of urban driving and human-object interactions.
The results in \Cref{tab:ablation-results} (middle) show that the model trained on these smaller datasets performs poorly across all tests.
This evidence supports one of our claims, that large-scale and diverse video data is crucial for training a generalist world model.

\myparagraph{Visual encoder.}
Since we do not reconstruct pixels, but latent features, the pre-trained DINOv2 encoder is crucial to the ability of the model to understand the world.
To validate our choice, we train two additional models with different encoders:
the VAE of Stable Diffusion 3.5 \citep{esser2024scalingrectifiedflowtransformers}, a small 84M-parameter model optimized for image compression and reconstruction fidelity,
and SigLIP~2 SO400M \citep{tschannen2025siglip2multilingualvisionlanguage}, a larger model trained for vision-language alignment.
Since the VAE operates on \(8{\times}8\) patches with a 16-dimensional embedding, we concatenate the features of four neighboring patches to reduce the number of tokens and increase the embedding dimension.
The results in \Cref{tab:ablation-results} (right) confirm DINOv2 as the best choice for representing frame features.
The VAE performs poorly on all tasks, as it is not designed for image understanding and would likely require a much larger predictor as in COSMOS.
SigLIP2 performs slightly worse than DINOv2, which we attribute to the noisy features derived from its vision-language pre-training.

\begin{table}[b]
  \centering
  \vspace{-1em}
  \caption{
    \textbf{Ablation studies.}
    For each setup, we report physical understanding on IntPhys, and mid-term segmentation on Cityscapes and VSPW.
    \textbf{Predictor size (left):}
    We train a base (86M), large (304M), and giant (1.1B) predictor to assess scaling trends \wrt model capacity.
    \textbf{Training data (center):}
    We compare the smaller and more specialized datasets Cityscapes (CS) and Something-Something V2 (SSv2) \versus our large-scale collection of web videos.
    \textbf{Visual encoder (right):}
    We compare encoders of different sizes trained for pixel reconstruction (VAE), vision-language alignment (SigLIP2), or representation learning (DINOv2).
  }
  \label{tab:ablation-results}
  \vspace{0.7em}
  \begin{subtable}[t]{0.32\textwidth}
    \small
    \centering
    \begin{tabular}{@{}lc@{}c@{}c@{}}
        \toprule
        Predictor & IntPhys & CS & VSPW \\
        \midrule
        Base  & 84.9 & 47.7 & 45.4 \\
        Large & 89.1 & 51.9 & 46.4 \\
        Giant & 90.6 & 53.2 & 46.8 \\
        \bottomrule
    \end{tabular}
  \end{subtable}
  \hfill
  \begin{subtable}[t]{0.32\textwidth}
    \small
    \centering
    \begin{tabular}{@{}lc@{}c@{}c@{}}
        \toprule
        Dataset & IntPhys & CS & VSPW \\
        \midrule
        CS      & 66.7 & 45.6 & 23.1 \\
        SSv2    & 79.3 & 44.9 & 45.2 \\
        Ours 66M & 90.6 & 53.2 & 46.8 \\
        \bottomrule
    \end{tabular}
    \end{subtable}
  \hfill
  \begin{subtable}[t]{0.32\textwidth}
    \small
    \centering
    \begin{tabular}{@{}lc@{}c@{}c@{}}  %
        \toprule
        Encoder & IntPhys & CS & VSPW \\
        \midrule
        SD3.5 VAE & -- & 13.0 & \phantom{0}1.5 \\
        SigLIP2   & 80.7 & 50.5 & 41.0 \\
        DINOv2 & 90.6 & 53.2 & 46.8 \\
        \bottomrule
    \end{tabular}
  \end{subtable}  
\end{table}

\begin{figure*}[t]
    \centering
    \includegraphics[width=1\linewidth]{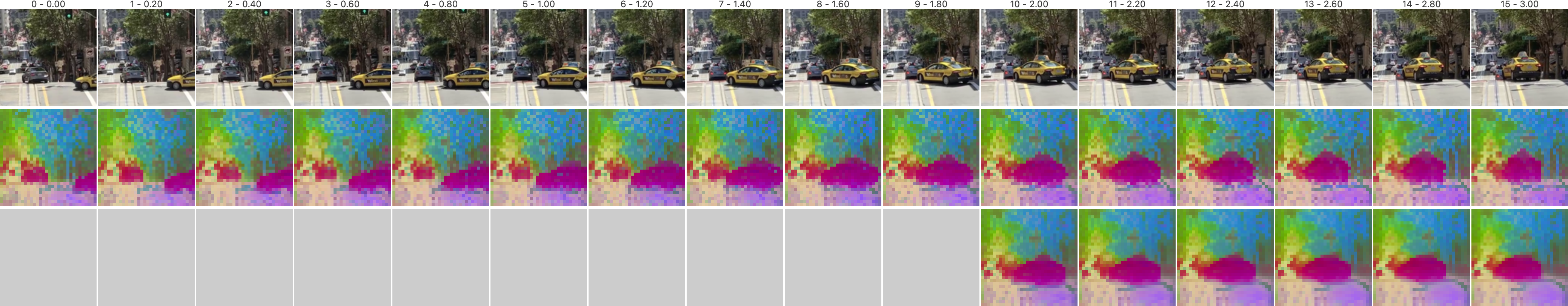}\vspace{0.4em}
    \includegraphics[width=1\linewidth]{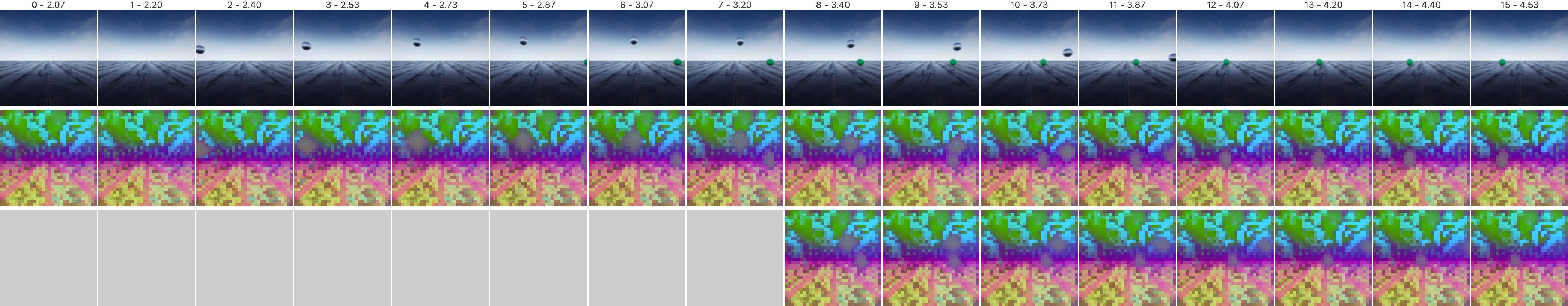}
    \caption{
        \textbf{Autoregressive predictions.}
        For each video, from top to bottom: frames with timestamps, encoder features, autoregressive predictions in latent space.
        The predictor has access to ``ground-truth'' encoder features for the first 8--10 frames (mid row, above the gray squares), and then to its own predictions (bottom row, past the gray squares).
        Latent features are visualized through a PCA projection computed on the encoder outputs.
    }
    \label{fig:feature-visualization}
    \vspace{-1em}
\end{figure*}

\myparagraph{Qualitative analysis.}
In \Cref{fig:feature-visualization}, we visualize the latent features predicted by our model.
For each video, we show the RGB frames, \ie the observations, and the corresponding features of the encoder, \ie the state.
The predictor is given a few initial frames and is asked to continue the sequence autoregressively: each new frame is predicted at once, and is fed back to the model as context in place of the encoder features.
The model correctly tracks the taxi turning at the crossroad, demonstrating common-sense understanding of traffic.
And in the IntPhys video, it correctly tracks the linear and parabolic motions of the objects.
In both cases, the longer the temporal horizon, the higher the uncertainty, and the more the predictions become blurry.
Finally, these examples showcase the ability of taking in a context of varying length, and to predict features at different frame rates.

\begin{figure}
    \centering
    \begin{minipage}[b][2.7in]{.49\textwidth}
        \centering
        \includegraphics[width=1\linewidth]{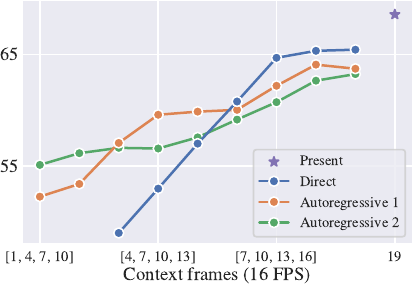}
        \captionof{figure}{
            \textbf{How far can the model predict?}
            Cityscapes segmentation forecasting performance as context frames are progressively shifted back in time, further away from the target frame 19.
            We compare direct prediction \versus 1- or 2-step autoregressive prediction.
        }
        \label{fig:autoregressive}
        \label{fig:test1}
    \end{minipage}\hfill%
    \begin{minipage}[b][2.7in]{.45\textwidth}
        \centering
        \includegraphics[width=0.31\linewidth,cfbox=black 0.5pt 0pt 0pt]{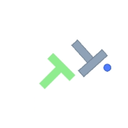}\hfill
        \includegraphics[width=0.31\linewidth,cfbox=black 0.5pt 0pt 0pt]{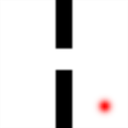}\hfill
        \includegraphics[width=0.31\linewidth,cfbox=black 0.5pt 0pt 0pt]{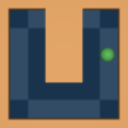}\vspace{1em}
        \begin{tabular}{@{}l ccc@{}}
            \toprule
            Model & PushT & Wall & PointMaze \\
            \midrule
            Scratch     & 46.9 & 87.1 & 59.4 \\
            Action-only & 49.4 & 91.1 & 61.6 \\
            Fine-tuned  & 59.4 & 93.8 & 68.7 \\
            \bottomrule
        \end{tabular}\vspace{1em}
        \captionof{table}{
            \textbf{Planning evaluations.}
            Success rate of a planner that refines candidate trajectories by rolling them out in the latent space of our action-conditioned world model.        
            \phantom{conditioned world model. conditioned world model.}
        }
        \label{tab:planning-evals}
    \end{minipage}
    \vspace{-1em}
\end{figure}

\myparagraph{Direct \versus auto-regressive prediction.}
During training, the model is exposed to a specific range of time deltas between frames (\Cref{sec:method-world-model-architecture}).
How far in the future can the model predict, generalizing beyond its training distribution and taking guesses about the inherent uncertainty of the world?
Since our model can operate with arbitrary query timestamps, when the prediction target is far from the last observed frame, we can take two approaches:
either make a \emph{direct prediction}, \ie query the model directly with the target timestamp, or predict intermediate frames \emph{autoregressively}, breaking down the interval into smaller steps.
Using Cityscapes as a benchmark, we compare the two approaches by evaluating segmentation forecasting as the context frames are progressively shifted back and away from the target frame.
From the results in \Cref{fig:autoregressive}, we observe that direct predictions are more accurate at short time deltas, while the autoregressive rollout holds better at longer horizons.
However, all predictions become inaccurate as the forecasting interval approaches 1 second.
Forecasting at longer horizons remains a limitation of current models.

\subsection{Action-conditioned fine-tuning and planning evaluations}
\label{sec:experiments-action-conditioning}

As a final experiment, we post-train our world model on action-conditioned data, and evaluate its performance on planning tasks with the setup of \citet{zhou2024dinowm}.
First, a world model is trained on offline trajectories, \ie observations and action pairs \(\Set{(\vv_t, \va_t)}_{t=1}^T\), for each of three environments: PushT \citep{chi2023diffusion}, Wall \citep{zhou2024dinowm}, and PointMaze \citep{fu2020d4rl}.
Then, a planning algorithm uses the learned model to roll out candidate trajectories and iteratively assess the distance from the goal, until a final plan is chosen for execution in the real environment.
Details about the environments, offline trajectories, and planning are summarized in \Cref{appendix:planning-evals}.

As detailed in \cref{sec:method-actions}, we do not need to train a predictor from scratch for each environment.
Instead, we can leverage the large-scale pre-training and add \emph{action blocks} to the architecture.
For each environment, we train the action-conditioned models on offline trajectories for 25 epochs, with clips of \(T=4\) frames at 224 pixels, starting from the pre-trained ``base'' model.
As a point of comparison, we also train a model where all parameters are frozen except the action blocks, and another model initialized from scratch.
In \Cref{tab:planning-evals}, we report the success rate over 512 episodes per environment.
The main observation is that large-scale pre-training has a positive effect on performance, compared to training the predictor from scratch.
We expect this benefit to be more pronounced in more complex environments that more closely resemble the pre-training data.

\section{Conclusion}

We presented DINO-world, a latent-space world model trained at scale on a large uncurated video dataset, filling a gap between massive pixel-space generative models~\citep{brooks2024video,agarwal2025cosmos,wang2025wan} and small-scale latent-space world models~\citep{ha2018world,Hafner2019DreamTC,karypidis2024dinoforesight}.
To best utilize the capacity of the predictor for learning temporal dynamics, our approach leverages a frozen vision encoder with strong semantic features, namely DINOv2.
By design, the predictor is not constrained to a specific resolution, frame rate or context length, and can be easily adapted for action-conditioned post-training.
Furthermore, we conducted extensive evaluations, bringing together and comparing several approaches to world modeling.
The results on dense forecasting~\citep{luc2017predsemantic} and intuitive physics~\citep{garrido2025intuitive} demonstrate the benefits of latent-space world models, as well as the importance of large-scale pre-training.
Future directions include improving long-term predictions, \eg by sampling one of the possible futures, exploring strategies for data curation, validating post-training and planning in real-world environments, and incorporating language as the conditioning signal.

\FloatBarrier
\clearpage

{
    \small
    \bibliographystyle{unsrtnat}
    \bibliography{main}

\begin{thebibliography}{69}
\providecommand{\natexlab}[1]{#1}
\providecommand{\url}[1]{\texttt{#1}}
\expandafter\ifx\csname urlstyle\endcsname\relax
  \providecommand{\doi}[1]{doi: #1}\else
  \providecommand{\doi}{doi: \begingroup \urlstyle{rm}\Url}\fi

\bibitem[Ha and Schmidhuber(2018)]{ha2018world}
David Ha and J{\"u}rgen Schmidhuber.
\newblock Recurrent world models facilitate policy evolution.
\newblock In \emph{NeurIPS}, 2018.

\bibitem[Hu et~al.(2023)Hu, Russell, Yeo, Murez, Fedoseev, Kendall, Shotton,
  and Corrado]{hu2023gaia}
Anthony Hu, Lloyd Russell, Hudson Yeo, Zak Murez, George Fedoseev, Alex
  Kendall, Jamie Shotton, and Gianluca Corrado.
\newblock Gaia-1: A generative world model for autonomous driving.
\newblock \emph{arXiv preprint arXiv:2309.17080}, 2023.

\bibitem[Yang et~al.(2023)Yang, Du, Ghasemipour, Tompson, Schuurmans, and
  Abbeel]{yang2023learning}
Mengjiao Yang, Yilun Du, Kamyar Ghasemipour, Jonathan Tompson, Dale Schuurmans,
  and Pieter Abbeel.
\newblock Learning interactive real-world simulators.
\newblock In \emph{ICLR}, 2023.

\bibitem[Brooks et~al.(2024)Brooks, Peebles, Holmes, DePue, Guo, Jing, Schnurr,
  Taylor, Luhman, Luhman, Ng, Wang, and Ramesh]{brooks2024video}
Tim Brooks, Bill Peebles, Connor Holmes, Will DePue, Yufei Guo, Li~Jing, David
  Schnurr, Joe Taylor, Troy Luhman, Eric Luhman, Clarence Ng, Ricky Wang, and
  Aditya Ramesh.
\newblock Video generation models as world simulators.
\newblock Blog post, 2024.
\newblock URL
  \url{https://openai.com/research/video-generation-models-as-world-simulators}.

\bibitem[Bruce et~al.(2024)Bruce, Dennis, Edwards, Parker-Holder, Shi, Hughes,
  Lai, Mavalankar, Steigerwald, Apps, et~al.]{bruce2024genie}
Jake Bruce, Michael~D Dennis, Ashley Edwards, Jack Parker-Holder, Yuge Shi,
  Edward Hughes, Matthew Lai, Aditi Mavalankar, Richie Steigerwald, Chris Apps,
  et~al.
\newblock Genie: Generative interactive environments.
\newblock In \emph{ICML}, 2024.

\bibitem[Parker-Holder et~al.(2024)Parker-Holder, Ball, Bruce, Dasagi,
  Holsheimer, Kaplanis, Moufarek, Scully, Shar, Shi, Spencer, Yung, Dennis,
  Kenjeyev, Long, Mnih, Chan, Gazeau, Li, Pardo, Wang, Zhang, Besse, Harley,
  Mitenkova, Wang, Clune, Hassabis, Hadsell, Bolton, Singh, and
  Rockt{\"a}schel]{parkerholder2024genie2}
Jack Parker-Holder, Philip Ball, Jake Bruce, Vibhavari Dasagi, Kristian
  Holsheimer, Christos Kaplanis, Alexandre Moufarek, Guy Scully, Jeremy Shar,
  Jimmy Shi, Stephen Spencer, Jessica Yung, Michael Dennis, Sultan Kenjeyev,
  Shangbang Long, Vlad Mnih, Harris Chan, Maxime Gazeau, Bonnie Li, Fabio
  Pardo, Luyu Wang, Lei Zhang, Frederic Besse, Tim Harley, Anna Mitenkova, Jane
  Wang, Jeff Clune, Demis Hassabis, Raia Hadsell, Adrian Bolton, Satinder
  Singh, and Tim Rockt{\"a}schel.
\newblock Genie 2: A large-scale foundation world model.
\newblock Blog post, 2024.
\newblock URL
  \url{https://deepmind.google/discover/blog/genie-2-a-large-scale-foundation-world-model/}.

\bibitem[Bartoccioni et~al.(2025)Bartoccioni, Ramzi, Besnier, Venkataramanan,
  Vu, Xu, Chambon, Gidaris, Odabas, Hurych, Marlet, Boulch, Chen, Éloi
  Zablocki, Bursuc, Valle, and Cord]{bartoccioni2025vavam}
Florent Bartoccioni, Elias Ramzi, Victor Besnier, Shashanka Venkataramanan,
  Tuan-Hung Vu, Yihong Xu, Loick Chambon, Spyros Gidaris, Serkan Odabas, David
  Hurych, Renaud Marlet, Alexandre Boulch, Mickael Chen, Éloi Zablocki, Andrei
  Bursuc, Eduardo Valle, and Matthieu Cord.
\newblock Vavim and vavam: Autonomous driving through video generative
  modeling.
\newblock \emph{arXiv preprint arXiv:2502.15672}, 2025.

\bibitem[Agarwal et~al.(2025)Agarwal, Ali, Bala, Balaji, Barker, Cai,
  Chattopadhyay, Chen, Cui, Ding, et~al.]{agarwal2025cosmos}
Niket Agarwal, Arslan Ali, Maciej Bala, Yogesh Balaji, Erik Barker, Tiffany
  Cai, Prithvijit Chattopadhyay, Yongxin Chen, Yin Cui, Yifan Ding, et~al.
\newblock Cosmos world foundation model platform for physical ai.
\newblock \emph{arXiv preprint arXiv:2501.03575}, 2025.

\bibitem[Russell et~al.(2025)Russell, Hu, Bertoni, Fedoseev, Shotton, Arani,
  and Corrado]{russell2025gaia2}
Lloyd Russell, Anthony Hu, Lorenzo Bertoni, George Fedoseev, Jamie Shotton,
  Elahe Arani, and Gianluca Corrado.
\newblock {GAIA-2}: A controllable multi-view generative world model for
  autonomous driving.
\newblock \emph{arXiv preprint arXiv:2503.20523}, 2025.
\newblock URL \url{https://arxiv.org/abs/2503.20523}.

\bibitem[Polyak et~al.(2025)Polyak, Zohar, Brown, Tjandra, Sinha, Lee, Vyas,
  Shi, Ma, Chuang, Yan, Choudhary, Wang, Sethi, Pang, Ma, Misra, Hou, Wang,
  Jagadeesh, Li, Zhang, Singh, Williamson, Le, Yu, Singh, Zhang, Vajda, Duval,
  Girdhar, Sumbaly, Rambhatla, Tsai, Azadi, Datta, Chen, Bell, Ramaswamy,
  Sheynin, Bhattacharya, Motwani, Xu, Li, Hou, Hsu, Yin, Dai, Taigman, Luo,
  Liu, Wu, Zhao, Kirstain, He, He, Pumarola, Thabet, Sanakoyeu, Mallya, Guo,
  Araya, Kerr, Wood, Liu, Peng, Vengertsev, Schonfeld, Blanchard, Juefei-Xu,
  Nord, Liang, Hoffman, Kohler, Fire, Sivakumar, Chen, Yu, Gao, Georgopoulos,
  Moritz, Sampson, Li, Parmeggiani, Fine, Fowler, Petrovic, and
  Du]{polyak2025moviegen}
Adam Polyak, Amit Zohar, Andrew Brown, Andros Tjandra, Animesh Sinha, Ann Lee,
  Apoorv Vyas, Bowen Shi, Chih-Yao Ma, Ching-Yao Chuang, David Yan, Dhruv
  Choudhary, Dingkang Wang, Geet Sethi, Guan Pang, Haoyu Ma, Ishan Misra,
  Ji~Hou, Jialiang Wang, Kiran Jagadeesh, Kunpeng Li, Luxin Zhang, Mannat
  Singh, Mary Williamson, Matt Le, Matthew Yu, Mitesh~Kumar Singh, Peizhao
  Zhang, Peter Vajda, Quentin Duval, Rohit Girdhar, Roshan Sumbaly, Sai~Saketh
  Rambhatla, Sam Tsai, Samaneh Azadi, Samyak Datta, Sanyuan Chen, Sean Bell,
  Sharadh Ramaswamy, Shelly Sheynin, Siddharth Bhattacharya, Simran Motwani,
  Tao Xu, Tianhe Li, Tingbo Hou, Wei-Ning Hsu, Xi~Yin, Xiaoliang Dai, Yaniv
  Taigman, Yaqiao Luo, Yen-Cheng Liu, Yi-Chiao Wu, Yue Zhao, Yuval Kirstain,
  Zecheng He, Zijian He, Albert Pumarola, Ali Thabet, Artsiom Sanakoyeu, Arun
  Mallya, Baishan Guo, Boris Araya, Breena Kerr, Carleigh Wood, Ce~Liu, Cen
  Peng, Dimitry Vengertsev, Edgar Schonfeld, Elliot Blanchard, Felix Juefei-Xu,
  Fraylie Nord, Jeff Liang, John Hoffman, Jonas Kohler, Kaolin Fire, Karthik
  Sivakumar, Lawrence Chen, Licheng Yu, Luya Gao, Markos Georgopoulos, Rashel
  Moritz, Sara~K. Sampson, Shikai Li, Simone Parmeggiani, Steve Fine, Tara
  Fowler, Vladan Petrovic, and Yuming Du.
\newblock Movie gen: A cast of media foundation models.
\newblock \emph{arXiv preprint arXiv:2410.13720}, 2025.
\newblock URL \url{https://arxiv.org/abs/2410.13720}.

\bibitem[Wang et~al.(2025)Wang, Ai, Wen, Mao, Xie, Chen, Yu, Zhao, Yang, Zeng,
  et~al.]{wang2025wan}
Ang Wang, Baole Ai, Bin Wen, Chaojie Mao, Chen-Wei Xie, Di~Chen, Feiwu Yu,
  Haiming Zhao, Jianxiao Yang, Jianyuan Zeng, et~al.
\newblock Wan: Open and advanced large-scale video generative models.
\newblock \emph{arXiv preprint arXiv:2503.20314}, 2025.

\bibitem[LeCun(2022)]{lecun2022path}
Yann LeCun.
\newblock A path towards autonomous machine intelligence, 2022.

\bibitem[Majumdar et~al.(2024)Majumdar, Ajay, Zhang, Putta, Yenamandra, Henaff,
  Silwal, Mcvay, Maksymets, Arnaud, Yadav, Li, Newman, Sharma, Berges, Zhang,
  Agrawal, Bisk, Batra, Kalakrishnan, Meier, Paxton, Sax, and
  Rajeswaran]{majumdar2024OpenEQA}
Arjun Majumdar, Anurag Ajay, Xiaohan Zhang, Pranav Putta, Sriram Yenamandra,
  Mikael Henaff, Sneha Silwal, Paul Mcvay, Oleksandr Maksymets, Sergio Arnaud,
  Karmesh Yadav, Qiyang Li, Ben Newman, Mohit Sharma, Vincent Berges, Shiqi
  Zhang, Pulkit Agrawal, Yonatan Bisk, Dhruv Batra, Mrinal Kalakrishnan,
  Franziska Meier, Chris Paxton, Alexander Sax, and Aravind Rajeswaran.
\newblock {OpenEQA}: Embodied question answering in the era of foundation
  models.
\newblock In \emph{CVPR}, pages 16488--16498, 2024.

\bibitem[Tong et~al.(2024)Tong, Liu, Zhai, Ma, LeCun, and
  Xie]{tong2024eyeswideshut}
Shengbang Tong, Zhuang Liu, Yuexiang Zhai, Yi~Ma, Yann LeCun, and Saining Xie.
\newblock Eyes wide shut? exploring the visual shortcomings of multimodal llms.
\newblock In \emph{CVPR}, pages 9568--9578, 2024.
\newblock \doi{10.1109/CVPR52733.2024.00914}.

\bibitem[Hafner et~al.(2025)Hafner, Pasukonis, Ba, and
  Lillicrap]{Hafner2023MasteringDD}
Danijar Hafner, Jurgis Pasukonis, Jimmy Ba, and Timothy Lillicrap.
\newblock Mastering diverse control tasks through world models.
\newblock \emph{Nature}, pages 1--7, 2025.

\bibitem[Alonso et~al.(2025)Alonso, Jelley, Micheli, Kanervisto, Storkey,
  Pearce, and Fleuret]{alonso2025diffusion}
Eloi Alonso, Adam Jelley, Vincent Micheli, Anssi Kanervisto, Amos~J Storkey,
  Tim Pearce, and Fran{\c{c}}ois Fleuret.
\newblock Diffusion for world modeling: Visual details matter in atari.
\newblock In \emph{NeurIPS}, 2025.

\bibitem[Hansen et~al.(2023)Hansen, Su, and Wang]{Hansen2023TDMPC2}
Nicklas Hansen, Hao Su, and Xiaolong Wang.
\newblock {TD-MPC2}: Scalable, robust world models for continuous control.
\newblock In \emph{ICLR}, 2023.

\bibitem[Zhou et~al.(2024)Zhou, Pan, LeCun, and Pinto]{zhou2024dinowm}
Gaoyue Zhou, Hengkai Pan, Yann LeCun, and Lerrel Pinto.
\newblock {DINO-WM}: World models on pre-trained visual features enable
  zero-shot planning.
\newblock \emph{arXiv preprint arXiv:2411.04983}, 2024.

\bibitem[Sobal et~al.(2025)Sobal, Zhang, Cho, Balestriero, Rudner, and
  LeCun]{sobal2025learning}
Vlad Sobal, Wancong Zhang, Kynghyun Cho, Randall Balestriero, Tim G.~J. Rudner,
  and Yann LeCun.
\newblock Learning from reward-free offline data: A case for planning with
  latent dynamics models.
\newblock \emph{arXiv preprint arXiv:2502.14819}, 2025.

\bibitem[Karypidis et~al.(2024)Karypidis, Kakogeorgiou, Gidaris, and
  Komodakis]{karypidis2024dinoforesight}
Efstathios Karypidis, Ioannis Kakogeorgiou, Spyros Gidaris, and Nikos
  Komodakis.
\newblock {DINO-Foresight}: Looking into the future with dino.
\newblock \emph{arXiv preprint arXiv:2412.11673}, 2024.

\bibitem[Garrido et~al.(2025)Garrido, Ballas, Assran, Bardes, Najman, Rabbat,
  Dupoux, and LeCun]{garrido2025intuitive}
Quentin Garrido, Nicolas Ballas, Mahmoud Assran, Adrien Bardes, Laurent Najman,
  Michael Rabbat, Emmanuel Dupoux, and Yann LeCun.
\newblock Intuitive physics understanding emerges from self-supervised
  pretraining on natural videos.
\newblock \emph{arXiv preprint arXiv:2502.11831}, 2025.

\bibitem[Motamed et~al.(2025)Motamed, Culp, Swersky, Jaini, and
  Geirhos]{motamed2025generative}
Saman Motamed, Laura Culp, Kevin Swersky, Priyank Jaini, and Robert Geirhos.
\newblock Do generative video models understand physical principles?
\newblock \emph{arXiv preprint arXiv:2501.09038}, 2025.

\bibitem[Oquab et~al.(2024)Oquab, Darcet, Moutakanni, Vo, Szafraniec, Khalidov,
  Fernandez, Haziza, Massa, El-Nouby, et~al.]{oquab2024dinov2}
Maxime Oquab, Timoth{\'e}e Darcet, Th{\'e}o Moutakanni, Huy Vo, Marc
  Szafraniec, Vasil Khalidov, Pierre Fernandez, Daniel Haziza, Francisco Massa,
  Alaaeldin El-Nouby, et~al.
\newblock Dinov2: Learning robust visual features without supervision.
\newblock \emph{TMLR}, 2024.

\bibitem[Bardes et~al.(2024)Bardes, Garrido, Ponce, Chen, Rabbat, LeCun,
  Assran, and Ballas]{bardes2024revisiting}
Adrien Bardes, Quentin Garrido, Jean Ponce, Xinlei Chen, Michael Rabbat, Yann
  LeCun, Mahmoud Assran, and Nicolas Ballas.
\newblock Revisiting feature prediction for learning visual representations
  from video.
\newblock \emph{TMLR}, 2024.

\bibitem[Sutton(1991)]{Sutton1991Dyna}
Richard~S. Sutton.
\newblock Dyna, an integrated architecture for learning, planning, and
  reacting.
\newblock \emph{SIGART Bull.}, 2\penalty0 (4), 1991.

\bibitem[Schmidhuber(1990)]{Schmidhuber1990MakingTheWorld}
J\"urgen Schmidhuber.
\newblock Making the world differentiable: on using self supervised fully
  recurrent neural networks for dynamic reinforcement learning and planning in
  non-stationary environments.
\newblock \emph{Forschungsberichte, TU Munich}, FKI 126 90, 1990.

\bibitem[Goodwin and Sin(1984)]{goodwin1984adaptive}
G.C. Goodwin and K.S. Sin.
\newblock \emph{Adaptive Filtering Prediction and Control}.
\newblock Information and Systems Sciences Series. Prentice-Hall, 1984.
\newblock ISBN 9780130040695.

\bibitem[Hafner et~al.(2019{\natexlab{a}})Hafner, Lillicrap, Ba, and
  Norouzi]{Hafner2019DreamTC}
Danijar Hafner, Timothy~P. Lillicrap, Jimmy Ba, and Mohammad Norouzi.
\newblock Dream to control: Learning behaviors by latent imagination.
\newblock In \emph{ICLR}, 2019{\natexlab{a}}.

\bibitem[Micheli et~al.(2023)Micheli, Alonso, and
  Fleuret]{micheli2023transformers}
Vincent Micheli, Eloi Alonso, and Fran{\c{c}}ois Fleuret.
\newblock Transformers are sample-efficient world models.
\newblock In \emph{ICLR}, 2023.
\newblock URL \url{https://openreview.net/forum?id=vhFu1Acb0xb}.

\bibitem[Hafner et~al.(2019{\natexlab{b}})Hafner, Lillicrap, Fischer, Villegas,
  Ha, Lee, and Davidson]{Hafner2019plannet}
Danijar Hafner, Timothy~P. Lillicrap, Ian~S. Fischer, Ruben Villegas, David~R
  Ha, Honglak Lee, and James Davidson.
\newblock Learning latent dynamics for planning from pixels.
\newblock In \emph{ICML}, 2019{\natexlab{b}}.

\bibitem[Bar et~al.(2024)Bar, Zhou, Tran, Darrell, and
  LeCun]{bar2024navigationworldmodels}
Amir Bar, Gaoyue Zhou, Danny Tran, Trevor Darrell, and Yann LeCun.
\newblock Navigation world models, 2024.
\newblock URL \url{https://arxiv.org/abs/2412.03572}.

\bibitem[Kalchbrenner et~al.(2016)Kalchbrenner, van~den Oord, Simonyan,
  Danihelka, Vinyals, Graves, and
  Kavukcuoglu]{kalchbrenner2016videopixelnetworks}
Nal Kalchbrenner, Aaron van~den Oord, Karen Simonyan, Ivo Danihelka, Oriol
  Vinyals, Alex Graves, and Koray Kavukcuoglu.
\newblock Video pixel networks.
\newblock In \emph{ICML}, 2016.

\bibitem[Babaeizadeh et~al.(2018)Babaeizadeh, Finn, Erhan, Campbell, and
  Levine]{babaeizadeh2018stochastic}
Mohammad Babaeizadeh, Chelsea Finn, Dumitru Erhan, Roy~H. Campbell, and Sergey
  Levine.
\newblock Stochastic variational video prediction.
\newblock In \emph{ICLR}, 2018.

\bibitem[Denton and Fergus(2018)]{denton2018stochasticvideogen}
Remi Denton and Rob Fergus.
\newblock Stochastic video generation with a learned prior.
\newblock In \emph{ICML}, 2018.

\bibitem[Yan et~al.(2021)Yan, Zhang, Abbeel, and Srinivas]{yan2021videogpt}
Wilson Yan, Yunzhi Zhang, Pieter Abbeel, and Aravind Srinivas.
\newblock {VideoGPT}: Video generation using vq-vae and transformers.
\newblock \emph{arXiv preprint arXiv:2104.10157}, 2021.

\bibitem[Yu et~al.(2023)Yu, Cheng, Sohn, Lezama, Zhang, Chang, Hauptmann, Yang,
  Hao, Essa, and Jiang]{yu2023magvit}
Lijun Yu, Yong Cheng, Kihyuk Sohn, José Lezama, Han Zhang, Huiwen Chang,
  Alexander~G. Hauptmann, Ming-Hsuan Yang, Yuan Hao, Irfan Essa, and Lu~Jiang.
\newblock {MAGVIT}: Masked generative video transformer.
\newblock In \emph{CVPR}, 2023.

\bibitem[Gupta et~al.(2023)Gupta, Yu, Sohn, Gu, Hahn, Li, Essa, Jiang, and
  Lezama]{Gupta2023PhotorealisticVG}
Agrim Gupta, Lijun Yu, Kihyuk Sohn, Xiuye Gu, Meera Hahn, Fei-Fei Li, Irfan
  Essa, Lu~Jiang, and Jos{\'e} Lezama.
\newblock Photorealistic video generation with diffusion models.
\newblock In \emph{ECCV}, 2023.

\bibitem[Villegas et~al.(2023)Villegas, Babaeizadeh, Kindermans, Moraldo,
  Zhang, Saffar, Castro, Kunze, and Erhan]{villegas2023phenaki}
Ruben Villegas, Mohammad Babaeizadeh, Pieter-Jan Kindermans, Hernan Moraldo,
  Han Zhang, Mohammad~Taghi Saffar, Santiago Castro, Julius Kunze, and Dumitru
  Erhan.
\newblock Phenaki: Variable length video generation from open domain textual
  descriptions.
\newblock In \emph{ICLR}, 2023.
\newblock URL \url{https://openreview.net/forum?id=vOEXS39nOF}.

\bibitem[Kondratyuk et~al.(2023)Kondratyuk, Yu, Gu, Lezama, Huang, Hornung,
  Adam, Akbari, Alon, Birodkar, Cheng, Chiu, Dillon, Essa, Gupta, Hahn, Hauth,
  Hendon, Martinez, Minnen, Ross, Schindler, Sirotenko, Sohn, Somandepalli,
  Wang, Yan, Yang, Yang, Seybold, and Jiang]{Kondratyuk2024VideoPoetAL}
D.~Kondratyuk, Lijun Yu, Xiuye Gu, Jos{\'e} Lezama, Jonathan Huang, Rachel
  Hornung, Hartwig Adam, Hassan Akbari, Yair Alon, Vighnesh Birodkar, Yong
  Cheng, Ming-Chang Chiu, Josh Dillon, Irfan Essa, Agrim Gupta, Meera Hahn,
  Anja Hauth, David Hendon, Alonso Martinez, David~C. Minnen, David~A. Ross,
  Grant Schindler, Mikhail Sirotenko, Kihyuk Sohn, Krishna Somandepalli,
  Huisheng Wang, Jimmy Yan, Ming Yang, Xuan Yang, Bryan Seybold, and Lu~Jiang.
\newblock Videopoet: A large language model for zero-shot video generation.
\newblock In \emph{ICML}, 2023.

\bibitem[Oh et~al.(2015)Oh, Guo, Lee, Lewis, and
  Singh]{oh2015actioncondvideopred}
Junhyuk Oh, Xiaoxiao Guo, Honglak Lee, Richard Lewis, and Satinder Singh.
\newblock Action-conditional video prediction using deep networks in atari
  games.
\newblock In \emph{NeurIPS}, 2015.

\bibitem[Chiappa et~al.(2017)Chiappa, Racaniere, Wierstra, and
  Mohamed]{chiappa2017recurrentenvsims}
Silvia Chiappa, Sébastien Racaniere, Daan Wierstra, and Shakir Mohamed.
\newblock Recurrent environment simulators.
\newblock In \emph{ICLR}, 2017.

\bibitem[Kang et~al.(2024)Kang, Yue, Lu, Lin, Zhao, Wang, Huang, and
  Feng]{Kang2024HowFar}
Bingyi Kang, Yang Yue, Rui Lu, Zhijie Lin, Yang Zhao, Kaixin Wang, Gao Huang,
  and Jiashi Feng.
\newblock How far is video generation from world model: A physical law
  perspective.
\newblock \emph{arXiv preprint arXiv:2411.02385}, 2024.

\bibitem[Luc et~al.(2017)Luc, Neverova, Couprie, Verbeek, and
  LeCun]{luc2017predsemantic}
Pauline Luc, Natalia Neverova, Camille Couprie, Jakob Verbeek, and Yann LeCun.
\newblock Predicting deeper into the future of semantic segmentation.
\newblock In \emph{ICCV}, 2017.

\bibitem[kuang Chiu et~al.(2019)kuang Chiu, Adeli, and
  Niebles]{chiu2019segmentingfuture}
Hsu kuang Chiu, Ehsan Adeli, and Juan~Carlos Niebles.
\newblock Segmenting the future.
\newblock In \emph{IROS}, 2019.

\bibitem[Luc et~al.(2018)Luc, Couprie, LeCun, and Verbeek]{luc2018predinstance}
Pauline Luc, Camille Couprie, Yann LeCun, and Jakob Verbeek.
\newblock Predicting future instance segmentation by forecasting convolutional
  features.
\newblock In \emph{ECCV}, 2018.

\bibitem[Vondrick et~al.(2016)Vondrick, Pirsiavash, and
  Torralba]{vondrick2016anticipating}
Carl Vondrick, Hamed Pirsiavash, and Antonio Torralba.
\newblock Anticipating visual representations from unlabeled video.
\newblock In \emph{CVPR}, 2016.

\bibitem[Girdhar and Grauman(2021)]{girdhar2021anticipativevideotransformer}
Rohit Girdhar and Kristen Grauman.
\newblock Anticipative video transformer.
\newblock In \emph{ICCV}, 2021.

\bibitem[Zhong et~al.(2022)Zhong, Schneider, Voit, Stiefelhagen, and
  Beyerer]{zhong2022anticipativefeaturefusiontransformer}
Zeyun Zhong, David Schneider, Michael Voit, Rainer Stiefelhagen, and Jürgen
  Beyerer.
\newblock Anticipative feature fusion transformer for multi-modal action
  anticipation.
\newblock In \emph{WACV}, 2022.

\bibitem[Kingma and Welling(2022)]{kingma2022vae}
Diederik~P Kingma and Max Welling.
\newblock Auto-encoding variational bayes.
\newblock In \emph{ICLR}, 2022.

\bibitem[Van Den~Oord et~al.(2017)Van Den~Oord, Vinyals, et~al.]{oord2018vqvae}
Aaron Van Den~Oord, Oriol Vinyals, et~al.
\newblock Neural discrete representation learning.
\newblock In \emph{NeurIPS}, 2017.

\bibitem[Tschannen et~al.(2025)Tschannen, Gritsenko, Wang, Naeem,
  Alabdulmohsin, Parthasarathy, Evans, Beyer, Xia, Mustafa,
  et~al.]{tschannen2025siglip2multilingualvisionlanguage}
Michael Tschannen, Alexey Gritsenko, Xiao Wang, Muhammad~Ferjad Naeem, Ibrahim
  Alabdulmohsin, Nikhil Parthasarathy, Talfan Evans, Lucas Beyer, Ye~Xia, Basil
  Mustafa, et~al.
\newblock Siglip 2: Multilingual vision-language encoders with improved
  semantic understanding, localization, and dense features, 2025.

\bibitem[Zhu et~al.(2024)Zhu, Wang, Zhao, Min, Deng, Dou, Wang, Shi, Wang,
  Zhang, You, Zhang, Zhao, Xiao, Zhao, Lu, and Huang]{zhu2024worldmodelsurvey}
Zheng Zhu, Xiaofeng Wang, Wangbo Zhao, Chen Min, Nianchen Deng, Min Dou, Yuqi
  Wang, Botian Shi, Kai Wang, Chi Zhang, Yang You, Zhaoxiang Zhang, Dawei Zhao,
  Liang Xiao, Jian Zhao, Jiwen Lu, and Guan Huang.
\newblock Is sora a world simulator? a comprehensive survey on general world
  models and beyond.
\newblock \emph{arXiv preprint arXiv:2405.03520}, 2024.

\bibitem[Vaswani et~al.(2017)Vaswani, Shazeer, Parmar, Uszkoreit, Jones, Gomez,
  Kaiser, and Polosukhin]{vaswani2017attention}
Ashish Vaswani, Noam Shazeer, Niki Parmar, Jakob Uszkoreit, Llion Jones,
  Aidan~N Gomez, {\L}ukasz Kaiser, and Illia Polosukhin.
\newblock Attention is all you need.
\newblock In \emph{NeurIPS}, 2017.

\bibitem[Fu et~al.(2024)Fu, Lian, Wang, Shi, Wang, Yala, Darrell, Efros, and
  Goldberg]{fu2024rethinking}
Letian Fu, Long Lian, Renhao Wang, Baifeng Shi, XuDong Wang, Adam Yala, Trevor
  Darrell, Alexei~A Efros, and Ken Goldberg.
\newblock Rethinking patch dependence for masked autoencoders.
\newblock In \emph{NeurIPS 2024 Workshop: Self-Supervised Learning-Theory and
  Practice}, 2024.

\bibitem[Su et~al.(2024)Su, Ahmed, Lu, Pan, Bo, and Liu]{su2024roformer}
Jianlin Su, Murtadha Ahmed, Yu~Lu, Shengfeng Pan, Wen Bo, and Yunfeng Liu.
\newblock Roformer: Enhanced transformer with rotary position embedding.
\newblock \emph{Neurocomputing}, 568:\penalty0 127063, 2024.

\bibitem[Darcet et~al.(2023)Darcet, Oquab, Mairal, and
  Bojanowski]{darcet2023registers}
Timoth{\'e}e Darcet, Maxime Oquab, Julien Mairal, and Piotr Bojanowski.
\newblock Vision transformers need registers.
\newblock \emph{TMLR}, 2023.

\bibitem[Loshchilov and Hutter(2019)]{loshchilov2019decoupled}
Ilya Loshchilov and Frank Hutter.
\newblock Decoupled weight decay regularization.
\newblock In \emph{ICLR}, 2019.

\bibitem[Cordts et~al.(2016)Cordts, Omran, Ramos, Rehfeld, Enzweiler, Benenson,
  Franke, Roth, and Schiele]{cordts2016cityscapes}
Marius Cordts, Mohamed Omran, Sebastian Ramos, Timo Rehfeld, Markus Enzweiler,
  Rodrigo Benenson, Uwe Franke, Stefan Roth, and Bernt Schiele.
\newblock The cityscapes dataset for semantic urban scene understanding.
\newblock In \emph{CVPR}, 2016.

\bibitem[Goyal et~al.(2017)Goyal, Ebrahimi~Kahou, Michalski, Materzynska,
  Westphal, Kim, Haenel, Fruend, Yianilos, Mueller-Freitag,
  et~al.]{goyal2017something}
Raghav Goyal, Samira Ebrahimi~Kahou, Vincent Michalski, Joanna Materzynska,
  Susanne Westphal, Heuna Kim, Valentin Haenel, Ingo Fruend, Peter Yianilos,
  Moritz Mueller-Freitag, et~al.
\newblock The" something something" video database for learning and evaluating
  visual common sense.
\newblock In \emph{ICCV}, 2017.

\bibitem[Miech et~al.(2019)Miech, Zhukov, Alayrac, Tapaswi, Laptev, and
  Sivic]{Miech2019HowTo100MLA}
Antoine Miech, Dimitri Zhukov, Jean-Baptiste Alayrac, Makarand Tapaswi, Ivan
  Laptev, and Josef Sivic.
\newblock {HowTo100M}: Learning a text-video embedding by watching hundred
  million narrated video clips.
\newblock In \emph{ICCV}, pages 2630--2640, 2019.

\bibitem[Kay et~al.(2017)Kay, Carreira, Simonyan, Zhang, Hillier,
  Vijayanarasimhan, Viola, Green, Back, Natsev, et~al.]{kay2017kinetics}
Will Kay, Joao Carreira, Karen Simonyan, Brian Zhang, Chloe Hillier, Sudheendra
  Vijayanarasimhan, Fabio Viola, Tim Green, Trevor Back, Paul Natsev, et~al.
\newblock The kinetics human action video dataset.
\newblock \emph{arXiv preprint arXiv:1705.06950}, 2017.

\bibitem[Miao et~al.(2021)Miao, Wei, Wu, Liang, Li, and Yang]{miao2021vspw}
Jiaxu Miao, Yunchao Wei, Yu~Wu, Chen Liang, Guangrui Li, and Yi~Yang.
\newblock Vspw: A large-scale dataset for video scene parsing in the wild.
\newblock In \emph{CVPR}, 2021.

\bibitem[Geiger et~al.(2013)Geiger, Lenz, Stiller, and
  Urtasun]{geiger2013vision}
Andreas Geiger, Philip Lenz, Christoph Stiller, and Raquel Urtasun.
\newblock Vision meets robotics: The kitti dataset.
\newblock \emph{The International Journal of Robotics Research}, 32\penalty0
  (11):\penalty0 1231--1237, 2013.

\bibitem[Riochet et~al.(2021)Riochet, Castro, Bernard, Lerer, Fergus, Izard,
  and Dupoux]{riochet2021intphys}
Ronan Riochet, Mario~Ynocente Castro, Mathieu Bernard, Adam Lerer, Rob Fergus,
  V{\'e}ronique Izard, and Emmanuel Dupoux.
\newblock Intphys 2019: A benchmark for visual intuitive physics understanding.
\newblock \emph{IEEE TPAMI}, 44\penalty0 (9):\penalty0 5016--5025, 2021.

\bibitem[Jassim et~al.(2024)Jassim, Holubar, Richter, Wolff, Ohmer, and
  Bruni]{jassim2024grasp}
Serwan Jassim, Mario Holubar, Annika Richter, Cornelius Wolff, Xenia Ohmer, and
  Elia Bruni.
\newblock Grasp: A novel benchmark for evaluating language grounding and
  situated physics understanding in multimodal language models.
\newblock In \emph{IJCAI}, 2024.

\bibitem[Weihs et~al.(2022)Weihs, Yuile, Baillargeon, Fisher, Marcus, Mottaghi,
  and Kembhavi]{weihs2022benchmarking}
Luca Weihs, Amanda Yuile, Ren{\'e}e Baillargeon, Cynthia Fisher, Gary Marcus,
  Roozbeh Mottaghi, and Aniruddha Kembhavi.
\newblock Benchmarking progress to infant-level physical reasoning in {AI}.
\newblock \emph{TMLR}, 2022.

\bibitem[Esser et~al.(2024)Esser, Kulal, Blattmann, Entezari, Müller, Saini,
  Levi, Lorenz, Sauer, Boesel, Podell, Dockhorn, English, Lacey, Goodwin,
  Marek, and Rombach]{esser2024scalingrectifiedflowtransformers}
Patrick Esser, Sumith Kulal, Andreas Blattmann, Rahim Entezari, Jonas Müller,
  Harry Saini, Yam Levi, Dominik Lorenz, Axel Sauer, Frederic Boesel, Dustin
  Podell, Tim Dockhorn, Zion English, Kyle Lacey, Alex Goodwin, Yannik Marek,
  and Robin Rombach.
\newblock Scaling rectified flow transformers for high-resolution image
  synthesis.
\newblock In \emph{ICML}, 2024.

\bibitem[Chi et~al.(2023)Chi, Xu, Feng, Cousineau, Du, Burchfiel, Tedrake, and
  Song]{chi2023diffusion}
Cheng Chi, Zhenjia Xu, Siyuan Feng, Eric Cousineau, Yilun Du, Benjamin
  Burchfiel, Russ Tedrake, and Shuran Song.
\newblock Diffusion policy: Visuomotor policy learning via action diffusion.
\newblock \emph{The International Journal of Robotics Research}, 2023.

\bibitem[Fu et~al.(2020)Fu, Kumar, Nachum, Tucker, and Levine]{fu2020d4rl}
Justin Fu, Aviral Kumar, Ofir Nachum, George Tucker, and Sergey Levine.
\newblock D4rl: Datasets for deep data-driven reinforcement learning.
\newblock \emph{arXiv preprint arXiv:2004.07219}, 2020.

\end{thebibliography}
}

\clearpage
\appendix
\section{Implementation and training}
\label{appendix:world-model-details}

\subsection{Architecture}
\label{appendix:world-model-architecture}

\paragraph{Video encoder.}
For our main experiments, we use a DINOv2 ViT-B/14 with registers \cite{oquab2024dinov2} as the frame encoder.
The model has 12 transformer blocks, embedding dimension 768, 12 attention heads, and a patch size of \(14{\times}14\).
We use the released code and weights from \href{https://github.com/facebookresearch/dinov2}{the official repository}.
We process each frame separately and take the output of the last transformer after the final \textsc{LayerNorm} as the frame representation.
Specifically, from an input frame of size \(224{\times}224\), we obtain \(256\) patch tokens with dimension \(768\).
The CLS token and the registers are discarded.

For the ablation studies using alternative image encoders, we use the following models:
\begin{itemize}
    \item The SigLIP 2 SO400M model \citep{tschannen2025siglip2multilingualvisionlanguage}, downloaded from \href{https://huggingface.co/google/siglip2-so400m-patch14-384}{Hugging Face Hub}.
    The model uses a patch size of \(14{\times}14\), same as DINOv2, and has a larger embedding dimension of 1152.
    \item The VAE tokenizer of Stable Diffusion 3.5 medium \citep{esser2024scalingrectifiedflowtransformers}, downloaded from \href{https://huggingface.co/stabilityai/stable-diffusion-3.5-medium}{Hugging Face Hub}.
    This model uses a smaller patch size of \(8{\times}8\) and smaller embedding dimension of 16. To better approximate the embedding dimension and sequence length of the other models, we concatenate the patch tokens of four adjacent patches, resulting in an embedding dimension of 64 and effective patch size of \(16{\times}16\).
\end{itemize}

\paragraph{Unconditional predictor.}

The predictor is a stack of pre-norm cross-attention transformer blocks that allow the query token(s) to attend to the context tokens produced by the video encoder.
At the beginning, the features of the video encoder are forwarded through a linear layer to match the embedding dimension of the transformer.
Similarly, at the end, the output of the last transformer block is passed through a \textsc{LayerNorm} and a linear layer to produce the final prediction in the same space as the video encoder.
\Cref{tab:predictor-sizes} summarizes the configurations of the transformer model used in our experiments.

Each token, whether it is a patch token from the video encoder used as context or a query token for future prediction, is associated with positional information.
The positional information is a tuple \((\tau, i, j)\) that represents the presentation timestamp in seconds and the spatial coordinates of the patch in the original video.
At every transformer block, we apply the rotation operation described in RoPE \citep{su2024roformer} to the tokens to inject the positional information in the attention operation.
Specifically, since we need to encode positional information along three axis, we split the 60 dimensions of the head in three 20-dimensional chunks, to which we apply RoPE rotations derived from 10 angular periods.
For each head, the trailing 4 dimensions of the embedding are left unchanged, \ie not rotated.

Typical values for the temporal positions are between 0 and 5 seconds, and between 0 and 1 for the normalized spatial position.
Therefore, we chose the ten angular periods for RoPE to be equally spaced between \(10^{-2}\) and \(10^2\), to cover a wide range of relative distances in time and space:
\begin{equation}
    \bm{\omega} = 10^{\texttt{linspace}(-2,\ 2,\ \texttt{steps}=10)}
\end{equation}

\paragraph{Action blocks.}
When adapting the unconditional predictor to an action-conditioned setting, we add \emph{action blocks} to the model, one after each transformer block.
At the beginning of the model, a linear layer projects the small-dimensional action representation to the same embedding dimension as the transformer.
Then, the action blocks take as input the query token and the action embedding, concatenate them, and pass them through the same operations of a regular residual MLP block whose D-dimensional output is added to the query token only.
At initialization, the layer scale parameters of the action blocks are set to zero, so they act as an identity.
Adding 12 action blocks to the ``base'' predictor, increases the parameter count by approximately 22M.

\subsection{Optimization}
\label{appendix:world-model-optimization}

\paragraph{Training objective.}
The output of the predictor is directly compared to the output of the video encoder for the corresponding frame.
As the training objective, we aim to minimize the element-wise smooth L1 loss between the predicted patch features and the target ones,
\(\forall i',j',t \in \Set{1, \ldots, T{-}1}\):
\begin{equation}
    \Loss(
        x_{t+1,i',j'},
        \hat{x}_{t+1,i',j'},
    ) = 
    \begin{cases}
        \frac{1}{2\beta} (x - \hat{x})^2, & \text{if } |x - \hat{x}| < \beta \\
        |x - \hat{x}| - \frac{\beta}{2}, & \text{otherwise}
    \end{cases}
    \qquad
    \text{with } \beta = 0.1.
\end{equation}

\paragraph{Hyperparameters.}
We train the predictor using the AdamW optimizer \citep{loshchilov2019decoupled} for 300k iterations with a batch size of 1024.
The learning rate is linearly increased from zero to \(10^{-4}\) in 5k iterations and then kept constant.
Weight decay is set to a constant value of \(0.4\).
We found no significant difference in performance when using a cosine-decayed learning rate schedule and/or a progressively increased weight decay.

\paragraph{Compute resources.}
We train all models in a multi-node setup, where each node contains eight NVIDIA H100 GPUs with 80GB of memory.
The training times and compute resources for the base, large, and giant predictors are summarized in \Cref{tab:predictor-sizes}.

\begin{table}[H]
\centering
\caption{
    \textbf{Model sizes and compute resources.}
    For each model size, we report the embedding dimension, number of transformer blocks, number of attention heads, MLP ratio, and total number of parameters.
    We also report the number of GPU nodes and training hours used to train the models.
    }
\label{tab:predictor-sizes}
\begin{tabular}{@{}l ccccc cc@{}}
\toprule
  & Embedding & Blocks & Heads & MLP ratio & Params. & Num. nodes & Training hours\\
\midrule
Base   & \phantom{0}768  & 12 & 12  & 4.0 & \phantom{0}86M  & \phantom{0}4 & 45.3 \\
Large  &           1024 & 24 & 16  & 4.0 & 304M & \phantom{0}8 & 70.2  \\
Giant  &           1536 & 40 & 24  & 4.0 & \phantom{0}1.1B &           16 & 95.6  \\
\bottomrule
\end{tabular}
\end{table}

\clearpage
\section{Datasets}
\label{appendix:datasets}

\begin{figure}[H]
    \centering
    \includegraphics[width=0.46\linewidth]{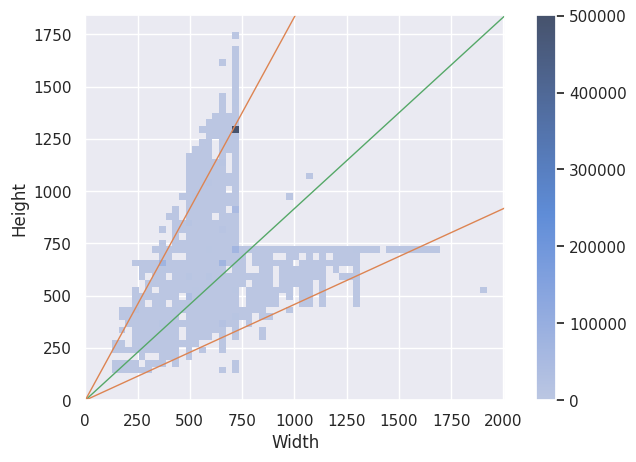}\hfill
    \includegraphics[width=0.525\linewidth]{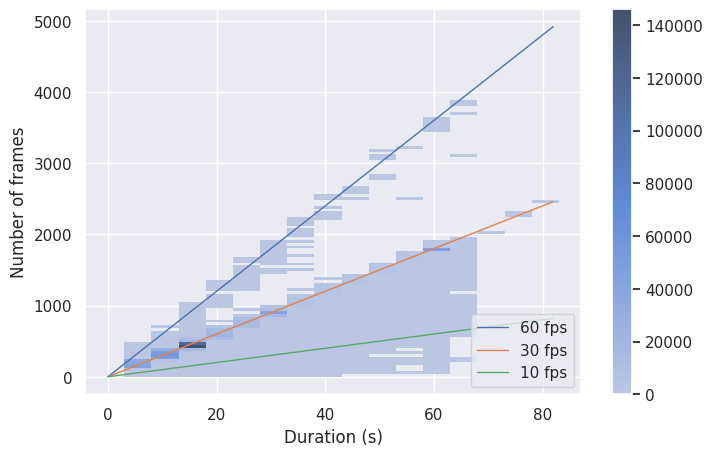}\hfill
    \caption{
        \textbf{Pre-training dataset statistics.}
        For our 66M video dataset, we report the joint histogram of height \versus width with highlighted aspect ratios \(16{:}9\), \(1{:}1\), and \(9{:}16\), as well as the distribution of number of frames \versus duration with highlighted 10, 30, 60 FPS.
    }
    \label{fig:dataset-stats}
\end{figure}

\paragraph{Pre-training datasets.}
To facilitate reproducing our results, we report key statistics of the unlabeled video dataset used to pre-train our video world model.
Specifically, \Cref{tab:dataset-stats} compares dataset size, duration FPS, and resolution of our dataset with Cityscapes \citep{cordts2016cityscapes} and Something-Something V2 (SSv2) \citep{goyal2017something}.
Furthermore, the histograms in \Cref{fig:dataset-stats} summarize the distribution of height \versus width (aspect ratios) and number of frames \versus duration for our dataset.
The content of the videos in our dataset is diverse and includes a wide range of activities, from cooking tutorials to outdoor scenes.

\begin{table}[H]
    \centering
    \caption{
        \textbf{Pre-training dataset statistics.}
        We report the number of videos, duration, frames per second (FPS), and resolution of the datasets used to pre-train our video world model.
    }
    \label{tab:dataset-stats}
    \begin{tabular}{@{}l lcccc@{}}
        \toprule
                   & Domain & Num. videos & Duration (s) & FPS & Resolution (\(H{\times}W\))\\
        \midrule
        Cityscapes & Driving & \phantom{00}2,975  & \({\sim}\)2 & 16 & \(1024{\times}2048\) \\
        SSv2       & Object manipulation & 168,913 & 2--6 & 12 & \(240{\times}240\) -- \(240{\times}463\) \\
        Ours       & General & \phantom{000}66M & 5--60 & 10--60 & Misc. \\
        \bottomrule
    \end{tabular}
\end{table}

\paragraph{Figures and assets.}

The taxi video in \Cref{fig:feature-visualization} (top) and \Cref{fig:appendix-feature-visualization} (bottom) is from the authors' private collection.
\Cref{fig:feature-visualization} (bottom) depicts a synthetic video from IntPhys \citep{garrido2025intuitive}, which is also used in \Cref{fig:appendix-feature-visualization} (bottom).
At the top of \Cref{tab:planning-evals}, the three images represent examples of the planning environments PushT \citep{chi2023diffusion}, Wall \citep{zhou2024dinowm}, and PointMaze \citep{fu2020d4rl}.
All other videos used for the visualizations in this paper are from a licensed video dataset.
In all cases, none of the videos used for the visualizations are part of the pre-training dataset.

\clearpage
\section{Unconditional world model evaluations}
\label{appendix:evaluation-details}

\subsection{Evaluation datasets}

\paragraph{Cityscapes.}
Dataset details:
\begin{itemize}
    \item Native FPS 16
    \item Resolution \(H{\times}W = 1024{\times}2048\) pixels, \(2{:}1\) aspect ratio
    \item Sequences of 30 frames, zero-indexed \([0, \ldots, 29]\)
    \item 19 segmentation classes
    \item For each video, only frame 19 is annotated, so each video is treated as a single sample for forecasting
    \item Forecasting protocol, for each video:
    \begin{itemize}
        \item Short-term prediction: take frames \([7, 10, 13, 16]\) and predict frame 19, \ie 187.5 ms in the future
        \item Mid-term prediction: take frames \([1, 4, 7, 10]\), predict frames \([13, 16, 19]\) autoregressively and evaluate frame 19, \ie 562.5 ms in the future
    \end{itemize}
\end{itemize}

\paragraph{VSPW.}
Dataset details:
\begin{itemize}
    \item Native FPS 15
    \item Resolution \(480{\times}854\) pixels, \(16{:}9\) aspect ratio
    \item The number of frames per video is not fixed, typically between 40 and 100 frames, zero-indexed
    \item All frames are annotated. From each video we consider all subsequences of 20 frames as samples for forecasting, \eg \([0, 1, \ldots, 19]\), \([1, 2, \ldots, 20]\), \([2, 3, \ldots, 21]\), etc.
    \item 124 segmentation classes
    \item Forecasting protocol, for each subsequence:
    \begin{itemize}
    \item Short-term prediction: take frames \([7, 10, 13, 16]\) and predict frame 19, \ie 200 ms in the future
    \item Mid-term prediction: take frames \([1, 4, 7, 10]\), predict frames \([13, 16, 19]\) autoregressively and evaluate frame 600 ms in the future
    \end{itemize}
\end{itemize}

\paragraph{KITTI.}
Dataset details:
\begin{itemize}
    \item Native FPS 10
    \item Resolution \(512{\times}1382\) pixels, cropped to \(352{\times}1216\) pixels (\(38{:}11\)) for monocular depth estimation
    \item KITTI videos contain vary between around 200 to 1000 frames, among which 652 are annotated in the test set, zero-indexed
    \item Forecasting protocol:
    \begin{itemize}
    \item Short-term prediction: take frames \([5, 7, 9, 11]\) and predict frame 13, \ie 200 ms in the future
    \item Mid-term prediction: take frames \([1, 3, 5, 7]\), predict frames \([9, 11, 13]\) autoregressively and evaluate frame 13, \ie 600 ms in the future
    \end{itemize}
\end{itemize}

\subsection{Baseline models}

\paragraph{COSMOS.}
The COSMOS model, specifically the autoregressive world model with diffusion decoder, has the following inference constraints:
\begin{itemize}
    \item The model operates at a fixed resolution of \(640{\times}1024\) pixels.
    \item The model can take as input a single frame or 9 frames, due to the implementation of the tokenizer.
    \item The model generates 24 frames at 25 FPS in an autoregressive manner.
\end{itemize}

\paragraph{DINO-Foresight.}
The DINO-Foresight model has the following inference constraints:
\begin{itemize}
    \item The model operates at a fixed resolution of \(448{\times}896\), because it was trained in two stages,
          first at low resolution \(224{\times}448\) and then at high resolution \(448{\times}896\).
    \item The model takes as input 4 frames and predicts the next frame in the sequence.
    \item Each frame is encoded with a frozen DINOv2 ViT-B/14 with registers, the features from the intermediate layers \([2, 5, 8, 11]\) (zero-indexed) are concatenated as a \(4 \cdot 768\) vector, and projected to the first 1152 principal components.
    \item After the predictor, the inverse PCA projection is applied, and the resulting features are split back into four 768-dimensional vectors.
\end{itemize}

\paragraph{V-JEPA.}
For the V-JEPA models, the following implementation details are important:
\begin{itemize}
    \item V-JEPA models contain two encoders, named \emph{encoder} and \emph{target encoder}.
    For the reasons explained below, \emph{both} are needed to run the intuitive physics and dense forecasting evaluations.
    At training time, the model is optimized as such:
        \begin{equation}
            \begin{aligned}
                \hat{\vx} &= \LayerNorm{\textsc{TargetEncoder}\mleft(\vv\mright)} \\
                \vx &= \Predictor{\Encoder{\vv}} \\
                \Loss &= \MSE{\vx}{\hat{\vx}}
            \end{aligned}
            \label{eq:vjepa-training}
        \end{equation}
        where the layer norm (LN) does not contain learnable parameters, and the gradient is only propagated to the encoder and predictor.
    Therefore, to extract present-time features, we run \(\LayerNorm{\textsc{TargetEncoder}\mleft(\vv\mright)}\), where the input image is duplicated to form a 2-frame video.
    Then, to predict features for a future frame given past frames, we run \(\Predictor{\Encoder{\vv}}\). 
    These features are assumed to be aligned with the output of the target encoder applied to the future frame.

    \item V-JEPA is trained at resolution \(224{\times}224\) with videos of \(T{=}16\) frames.
    However, it operates on 3D patches, or \emph{tubelets}, of size \(2{\times}16{\times}16\) pixels, and has a learned positional encoding of size \(8{\times}14{\times}14\).
    When we feed past frames to the model, they always need to be multiples of two.
        When the context has an odd number of frames, we drop the first frame to make it even.
    When the context frames result in a number of context tubelets that is shorter than the absolute positional encoding of the model, we slice the learned positional encoding and keep the first part.
            For example, 6 frames give 3 tubelets, so we use the first 3 entries out of 8.
    When the input frames have higher spatial resolution than the model, we bilinearly interpolate the positional encoding to match the input resolution. This operation does not affect the time dimension.
    Finally, to extract features for a specific frame in the future, we predict all tubelets up to that frame, rounding up if necessary, and then we select the corresponding tubelet from the output of the predictor.
\end{itemize}

\subsection{Evaluation protocols}

\paragraph{DINOv2 dense prediction heads.}
We train a linear head on top of DINOv2 ViT-B/14 with registers for each dense prediction task,
namely Cityscapes and VSPW semantic segmentation, and KITTI monocular depth estimation.

For semantic segmentation:
\begin{itemize}
    \item Training:
    \begin{enumerate}[label=\alph*)]
        \item We feed the model random crops at resolution \(448{\times}448\).
        \item The model applies one layer of batch normalization followed by a linear projection.
        \item We upsample the prediction logits to the original size using bilinear interpolation, and compute a cross-entropy loss with respect to the ground truth labels.
    \end{enumerate}
    \item Evaluation:
    \begin{enumerate}[label=\alph*)]
        \item We resize test images and labels such that the shortest side is 448 pixels.
        \item We apply the backbone and the head on \(448{\times}448\) crops using a sliding window with stride 300 pixels.
        \item The prediction logits of each crop are upsampled to the input size using bilinear interpolation.
        \item Finally, we average the prediction logits at each spatial location, and determine the predicted class with an \(\argmax\) operation.
    \end{enumerate}
\end{itemize}

For monocular depth estimation:
\begin{itemize}
    \item Training:
    \begin{enumerate}[label=\alph*)]
        \item We feed the model random crops at resolution \(352{\times}704\).
        \item The model applies one layer of batch normalization followed by a linear projection.
        \item We compute a mean squared error loss with respect to the ground truth depth labels.
    \end{enumerate}
    \item Evaluation:
    \begin{enumerate}[label=\alph*)]
        \item We apply the backbone and the head to whole images at resolution \(352{\times}1216\).
        \item The predicted depth values are upsampled to the input size using bilinear interpolation.
    \end{enumerate}
\end{itemize}

\paragraph{DINO-Foresight on Cityscapes.}
To evaluate DINO-Foresight on Cityscapes and VSPW, we train and test the segmentation heads at native model resolution of \(448{\times}896\), without sliding windows.
The predictor of DINO-Foresight takes as input the PCA-projected features of four intermediate DINOv2 ViT-B/14 layers, and is trained to predict future features in the same space.
Therefore, we can apply either of these two protocols:
\begin{itemize}
    \item Train the segmentation head on vanilla DINOv2 ViT-B features from the last layer. Then, at forecasting time, apply the inverse PCA projection to the output of the predictor, and slice the result to obtain features that are compatible with the segmentation head.
    \item Train the segmentation head on DINOv2 ViT-B features obtained by concatenating four intermediate layers, applying the PCA projection, applying the inverse projection, and slicing the result to obtain the 768-dimensional features of the last layer. Then, at inference time, we do the same with the output of the predictor.
\end{itemize}
The second protocol yields better results, likely thanks to the contribution of the intermediate layers for training the segmentation head, and is therefore the one used to report experimental results.

\paragraph{COSMOS on Cityscapes.}
To evaluate COSMOS on Cityscapes:
\begin{enumerate}[label=\alph*)]
    \item We resize Cityscapes videos to \(512{\times}1024\) and pad top and bottom with black bands to match \(640{\times}1024\).
    \item We consider every video as one sample for segmentation forecasting, since only frame 19 is annotated.
    \item For short-term predictions:
    \begin{itemize}
        \item We take the 9 frames at indices \([8, 9, 10, 11, 12, 13, 14, 15, 16]\), and feed them to COSMOS.
        \item COSMOS generates 24 frames corresponding to \([17, 18, 19, 20, \ldots, 40]\) in the original sequence.
        \item From the generated sequence, we take the 3rd frame, which corresponds to frame 19 in the original sequence.
    \end{itemize}
    \item For mid-term prediction:
    \begin{itemize}
        \item We take the 9 frames at indices \([2, 3, 4, 5, 6, 7, 8, 9, 10]\), and feed them to COSMOS.
        \item COSMOS generates 24 frames corresponding to \([11, 12, \ldots, 19, \ldots, 34]\) in the original sequence.
        \item From the generated sequence, we take the 9th frame, which corresponds to frame 19 in the original sequence.
    \end{itemize}
    \item We remove the padding and resize the image to \(448{\times}896\) to match the resolution of the pre-trained segmentation head.
    \item Finally, we apply the pre-trained DINO ViT-B backbone and the pre-trained segmentation head, and compare the output with the ground truth segmentation labels.
    \item In practice, both short-term and mid-term predictions are obtained in an autoregressive manner, since the model can not generate a frame at an arbitrary time step in the future,
\end{enumerate}

\paragraph{COSMOS on VSPW.}
To evaluate COSMOS on VSPW, we follow a similar procedure as for Cityscapes, except that:
\begin{enumerate}[label=\alph*)]
    \item We resize the longest side of the videos to 1024 pixels, and pad top and bottom with black bands.
    \item We consider every valid subsequence of length 20 as a sample for segmentation forecasting, since all frames are annotated. In other words, each video yields multiple samples.
    \item After removing the padding, we resize the frames such that the shortest side is 448 pixels, to match the resolution of the pre-trained segmentation head.
\end{enumerate}

\paragraph{COSMOS on KITTI.}
To evaluate COSMOS on KITTI, we follow a similar procedure as for Cityscapes and VSPW, except that after removing the padding, we resize the frames such that the shortest side is 352 pixels, to match the resolution of the pre-trained segmentation head.

\paragraph{DINO-Foresight on KITTI.}
To evaluate DINO-Foresight on KITTI, we interpolate the positional encoding so we are able to forward whole images at \(352{\times}1216\) resolution, as needed for KITTI.
For training the head, we apply the PCA projection and its inverse as done for Cityscapes.

\paragraph{V-JEPA on Cityscapes.}
To evaluate V-JEPA on Cityscapes, we train the head on layer-normed features of the target encoder.
Then, at inference time, we use the online encoder to produce features for the predictor and assume that the predictor output is aligned with the layer-normed features of the target encoder.
This protocol yields much better results than training the head on the online encoder.

\clearpage
\section{Action-conditioned fine-tuning and planning}
\label{appendix:planning-evals}

\subsection{Environments}
\label{appendix:planning-environments}

We use the offline trajectories released by \citet{zhou2024dinowm} for PushT, Wall and PointMaze to fine-tune our video model with action conditioning.
These trajectories are made of RGB images, actions, and proprioceptive observations.
In our setup, we ignore the proprioceptive observations and only use RGB images and actions.

\paragraph{Push-T.}
In this environment introduced by \citet{chi2023diffusion}, a pusher ball agent interacts with a T-shaped block.
Success is achieved when both the agent and the T-block, which start from a randomly initialized state, reach a target position.
The offline trajectories are stored at \(224{\times}224\) resolution.
The dataset provided in DINO-WM is made of 18500 samples, that are replays of the original released expert trajectories with various level of noise.

\paragraph{Point Maze.}
In this environment introduced by \citet{fu2020d4rl}, a force-actuated 2-DoF ball in the Cartesian directions $x$ and $y$ must reach a target position.
The agent's dynamics incorporate its velocity, acceleration, and inertia, making the movement realistic.
The offline trajectories are stored at \(224{\times}224\) resolution.
The Point Maze train set is made of 2000 fully random trajectories.

\paragraph{Wall.}
This 2D navigation environment introduced by \citet{zhou2024dinowm} features two rooms separated by a wall with a door. 
The agent's task is to navigate from a randomized starting location in one room to a goal in one of the two rooms, potentially passing through the door.
The offline trajectories are stored at \(224{\times}224\) resolution.
The Wall dataset is made 1920 random trajectories each with 50 time steps. 

\begin{table}[H]
    \centering
    \caption{
        Datasets used for action-conditioned fine-tuning and planning evaluations.
        We report the frame resolution, the total number of trajectories and the typical length of the trajectories.
        When using a ``frameskip'', the length of each trajectory is effectively divided by the number of skipped frames.
    }
    \label{tab:dinowm_hyperparams}
    \begin{tabular}{@{}lccc@{}}
        \toprule
                  & Resolution & Num. trajectories & Traj. length \\
        \midrule
        PointMaze          & 224 & \phantom{0}2000 & 100\\
        Push-T             & 224 & 18500 & 100-300\\
        Wall               & 224 & \phantom{0}1920 & 50\\
        \bottomrule
    \end{tabular}
\end{table}

\subsection{Action-conditioned fine-tuning.}
\label{appendix:action-conditioning-fine-tuning}

For the action-conditioned fine-tuning, we use the same training setup as for the unconditional models in terms of learning rate and batch size.
For each downstream environment, we train an action-conditioned model for the equivalent of 25 epochs using the offline trajectories.
A subset of 90\% of the trajectories, selected at random, is used for training, while the remaining 10\% is held out for the planning evaluation.
When training from scratch, the whole predictor is initialized from scratch and all parameters are trained.
When initializing from the video-only model, we either fine-tune the whole predictor or freeze all pre-trained parameters and only train the action blocks.
Following DINO-WM \citep{zhou2024dinowm}, we use a ``frameskip'' of 5 to speed up training and planning.
This means that for every RGB frame, we give the model the concatenation of the next 5 actions, and the model is tasked to predict the outcome.
Intermediate observations are not given to the model and not predicted by it, hence they are ``skipped'' and not ``stacked''.

\subsection{Planning algorithm}
\label{appendix:planning-algorithm}

\paragraph{Planning evaluation.}
We use the same planning setup as in DINO-WM \citep{zhou2024dinowm}.
We set a planning horizon $H$, encode the initial frame $\vx_t = \Encoder{\vv_t}$ and the goal frame $\vx_\text{goal} = \Encoder{\vv_\text{goal}}$, then optimize a sequence of actions $a_{t:t+H} := (a_t, \dots, a_{t+H-1})$ to minimize the planning objective \(L^p\) defined as the distance of the last predicted latent state to the goal state:
\begin{equation}
    L^p (
        \vv_t,
        a_{t:t+H},
        \vv_\text{goal}
    ) = \Norm{\hat{\vx}_{t+H} - \vx_\text{goal}}_{2},
\end{equation}
where each intermediate state is predicted from the previous as
\(\hat{\vx}_{i+1} = \Predictor{\hat{\vx}_i, a_i}\).
    
In our experiments, the initial and goal states are sampled from the validation set, such that the goal is reachable within $H{=}25$ steps from the initial state. 
The planning horizon of \(H{=}25\) means 25 individual actions, but only \(H'{=}5\) calls to the predictor due to the frameskip of \(f{=}5\).
Once the planner has determined the optimal trajectory, all \(25\) actions are stepped in the environment without replanning. 
The sequence of actions is optimized using the Cross-Entropy Method (CEM), described below.

\paragraph{Cross-Entropy Method.}
Planning at horizon $H$ is an optimization problem over the product  action space $\mathbb{R}^{H \times A}$, where each action is of dimension $A$.
Given an initial observation \(\vv_t\) and goal \(\vv_\text{goal}\), each action trajectory $a_{t:t+H}$ should be evaluated with a planning objective $L^p$.
The CEM optimisation algorithm proceeds as follows:
\begin{enumerate}[label=\alph*)]
    \item 
    At each iteration $j \in 1, \ldots, J$, the CEM samples a population of \( N \) action sequences, each of length \( H \), from a Gaussian distribution in $\Real^{H{\times}A}$.
    The Gaussian is initialized to have zero mean and unit variance, and will be updated iteratively by the CEM planner.
    \item 
    For each sampled action sequence \(a_{t:t+H}\), the world model is unrolled to predict the resulting state in the latent space:
    \begin{equation}
        \hat{\vx}_{i+1} = \Predictor{\hat{\vx}_i, a_i}, \qquad \forall i = t, \ldots, t+H-1.
    \end{equation}
    And the cost \( L^p(\vv_t, a_{t:t+H}, \vv_\text{goal}) \) is calculated for each candidate trajectory.
    \item 
    The top \( K \) action sequences with the lowest cost are selected, and the mean and covariance of the Gaussian proposal distribution are updated as the empirical mean and standard deviation of these $K$ action trajectories.
    \item 
    A new set of \( N \) action sequences is sampled from the updated distribution, and the process repeats for a fixed number of optimization steps $J$.
    \item 
    After $J$ iterations of optimization, the first $m$ actions \( (a_t, ...a_{t+m-1}) \), where $m$ is a planning hyperparameter, are executed in the environment.
\end{enumerate}

In \Cref{tab:planning_hyperparams}, we report the hyperparameters used to plan on each environment.
We keep the settings of DINO-WM~\cite{zhou2024dinowm} for Push-T, Wall and Maze, but reduce the  number of ``top'' actions, denoted $K$, to 10 instead of 30.
Note that when \(m = H\), the optimized sequence of actions is executed all at once and no replanning is done.

\begin{table}[H]
    \centering
    \caption{
        Environment-specific hyperparameters for planning:
        \(N\) is the number of action sequences evaluated in parallel in each planning iteration,
        \(H\) is the planning horizon,
        \(f\) is the frameskip,
        \(H'\) is the number of calls to the predictor,
        \(m\) is the number of actions stepped in the environment,
        \(K\) is the number of top trajectories selected at each iteration to update the proposal distribution, and
        \(J\) is the number of optimization iterations.
    }
    \label{tab:planning_hyperparams}
    \begin{tabular}{@{}lccccccc@{}}
        \toprule
                    & $N$ & $H$ & $f$ & $H'$ & $m$ & $K$ & $J$ \\
        \midrule
        Push-T      & 300 & 25 & 5 & 5 & 25 & 10 & 30 \\
        PointMaze   & 300 & 25 & 5 & 5 & 25 & 10 & 30 \\
        Wall        & 300 & 25 & 5 & 5 & 25 & 10 & 30 \\
        \bottomrule
    \end{tabular}
\end{table}

\clearpage
\section{Results}

\subsection{Qualitative visualizations}

\begin{figure}[H]
    \centering
    \includegraphics[width=1\linewidth]{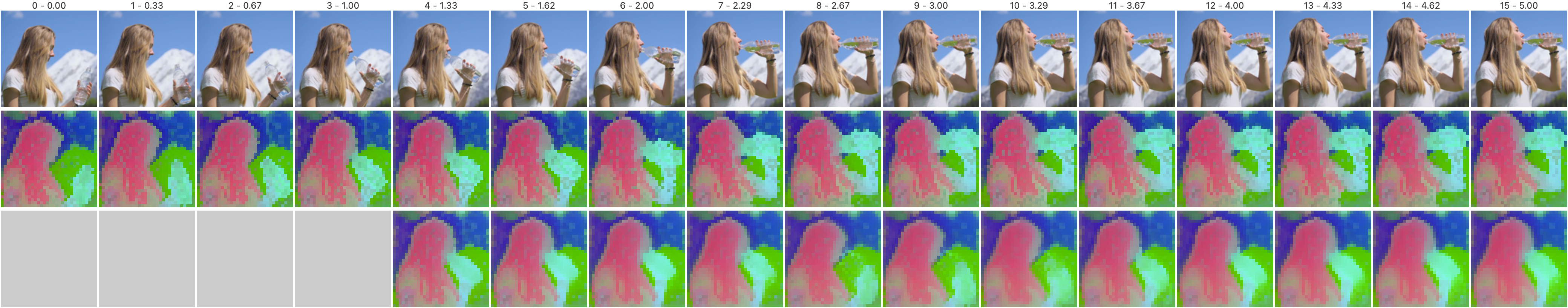}\vspace{0.5em}
    \includegraphics[width=1\linewidth]{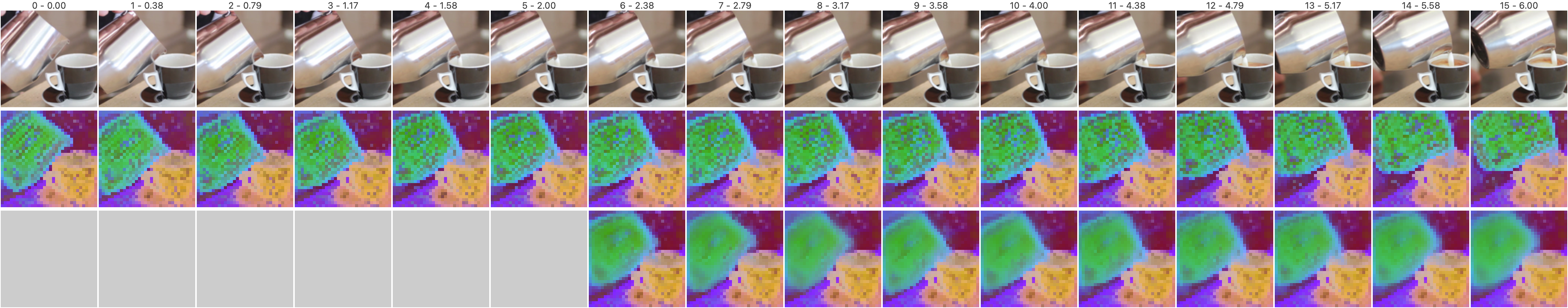}\vspace{0.5em}
    \includegraphics[width=1\linewidth]{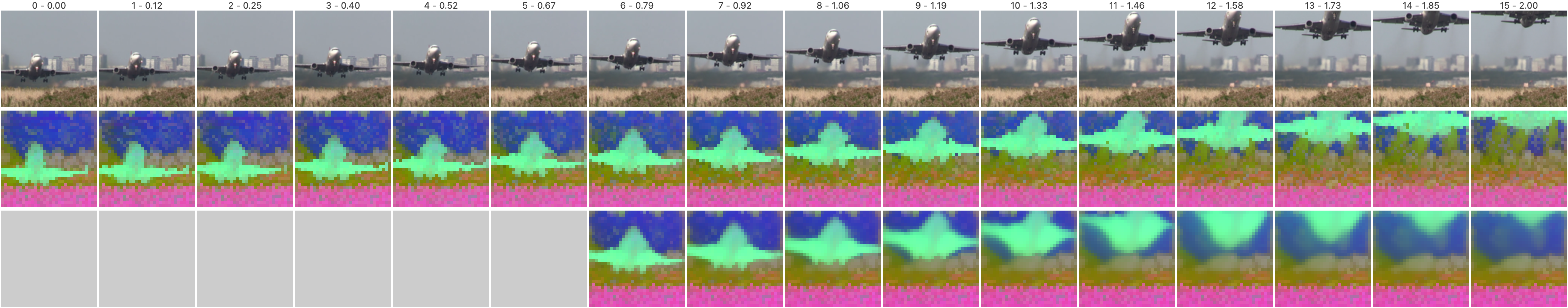}\vspace{0.5em}
    \includegraphics[width=1\linewidth]{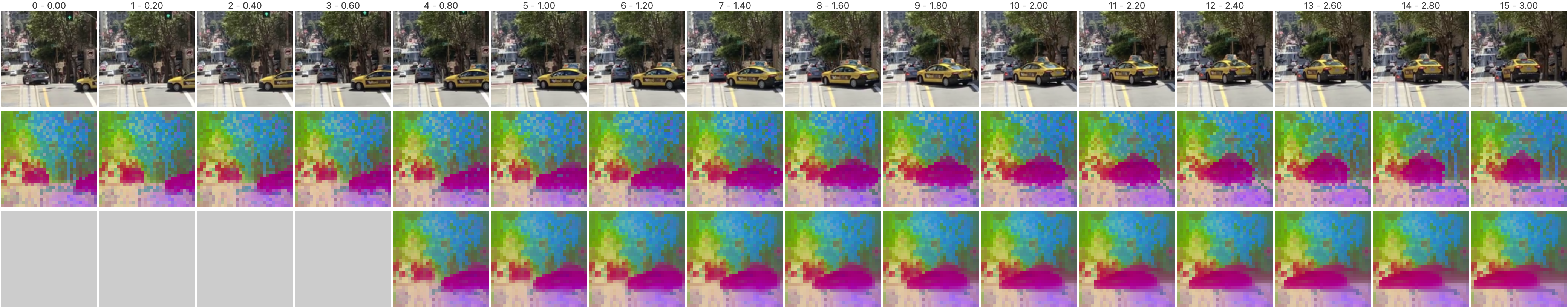}
    \caption{
    \textbf{Unconditional autoregressive rollouts in latent space.}
    For each clip, we feed the model a few initial frames, either 4 or 6, as processed by the encoder.
    We then roll out the predictor autoregressively, predicting one frame at a time and feeding it back to the predictor.
    From top to bottom, the three rows are: video frames with indices and timestamps, visualization of encoder features, visualization of autoregressive predictions of the world model.
    For each video, the predictor has access to the initial encoder features (middle row, all frames with a gray square below), and then to its own predictions (bottom row, past the gray squares).
    }
    \label{fig:appendix-feature-visualization}
\end{figure}

In \Cref{fig:appendix-feature-visualization}, we show additional visualizations of the world model predictions.
For each video, we feed the model a few initial frames as encoded by the encoder, and then we roll out the predictor autoregressively, predicting one frame at a time and feeding it back to the predictor.
Features are visualized as RGB by computing a 3-component PCA using the encoder features, and then applying the projection to both encoder features and predicted features.
In the first two clips, we highlight the tendency of the model to either predict plausible cyclical continuations or small localized movements.
In the third and fourth videos, we show how the predictions become progressively more blurry at longer prediction horizons where uncertainty about the future given the observed frames is higher.

\begin{figure}[H]
    \centering
    \begin{subfigure}[b]{1\textwidth}
         \centering
         \includegraphics[width=\linewidth]{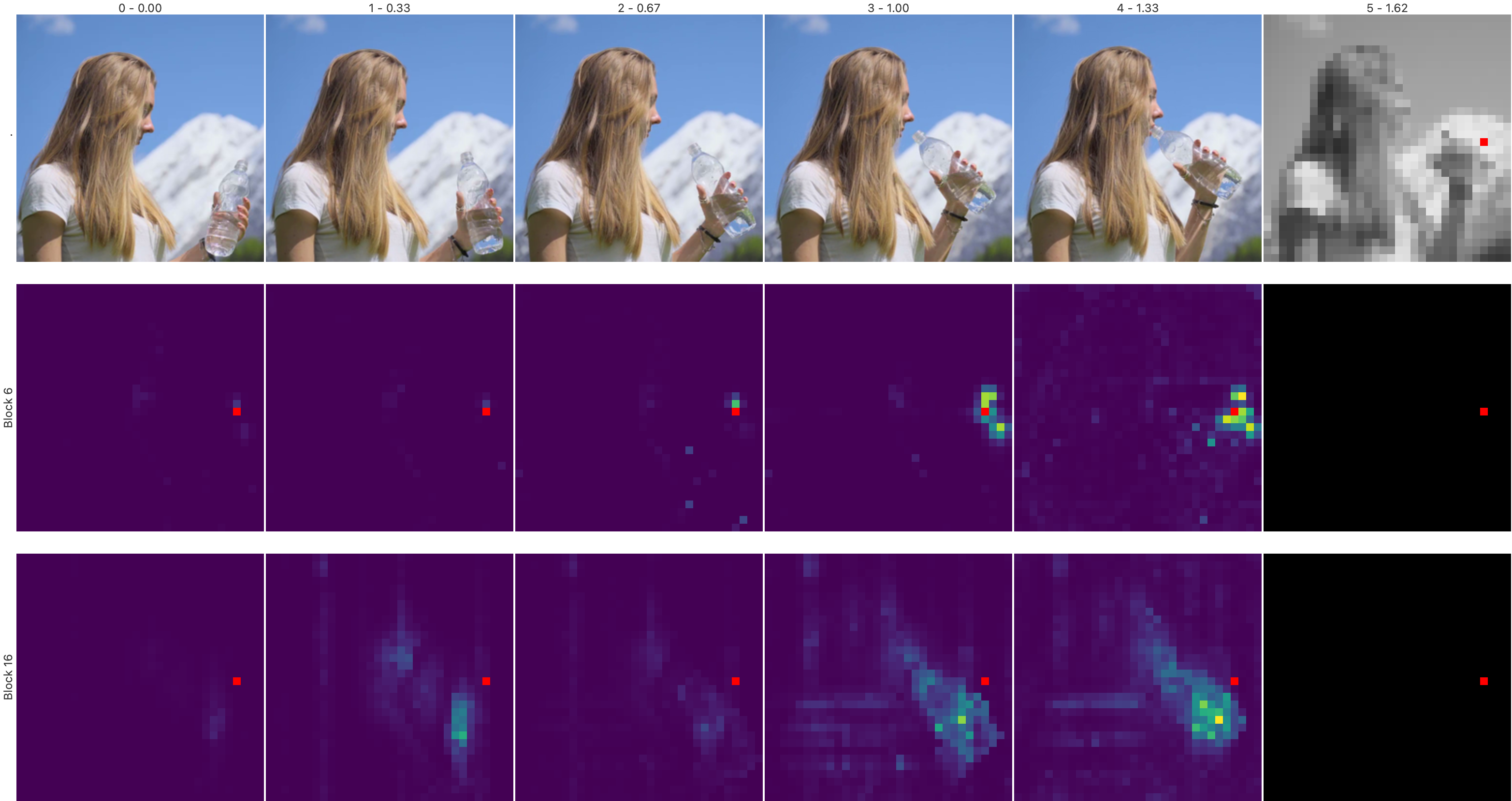}
     \end{subfigure}\vspace{1em}
     \begin{subfigure}[b]{1\textwidth}
         \centering
         \includegraphics[width=\linewidth]{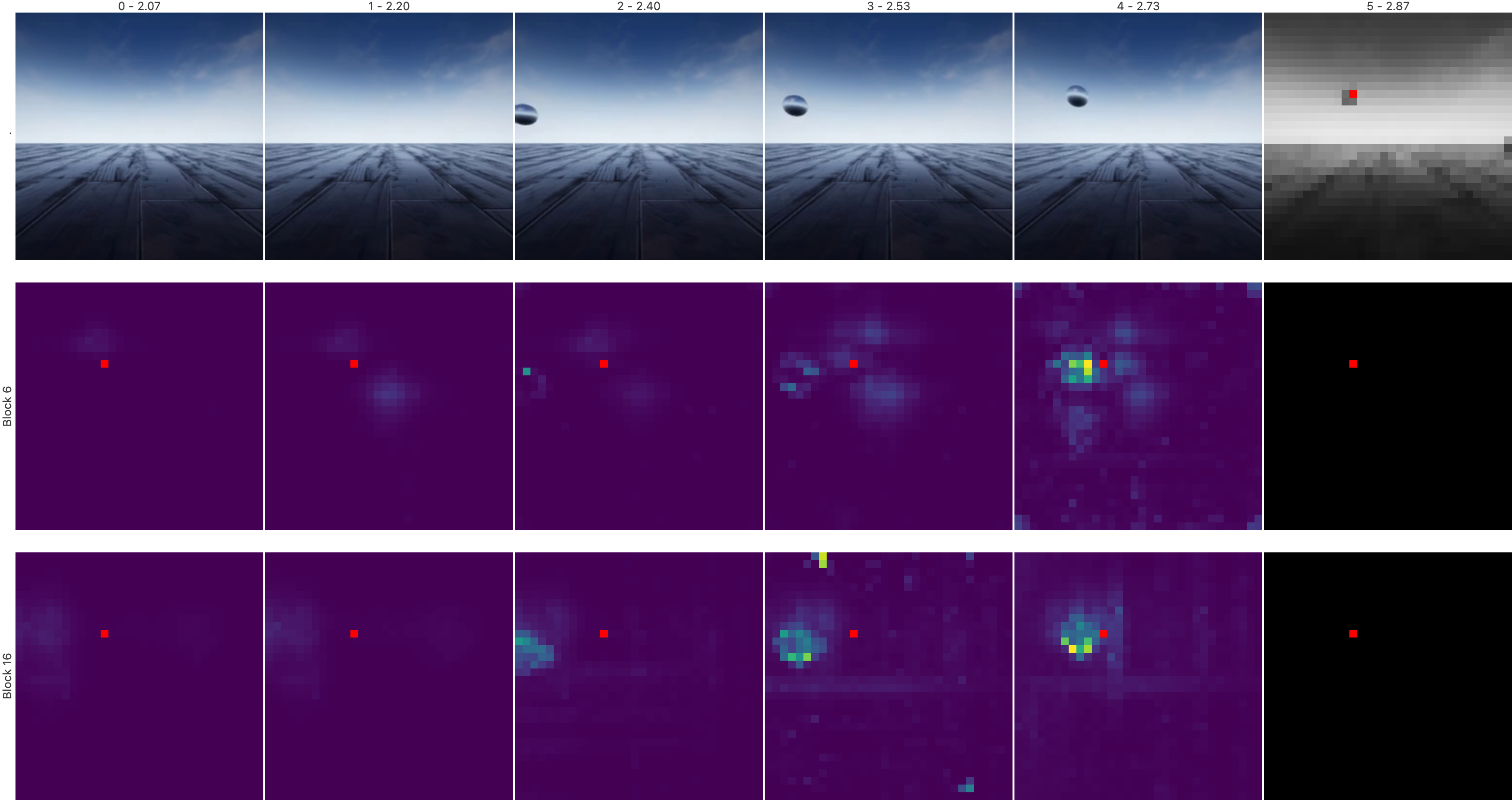}
     \end{subfigure}
    \caption{
    \textbf{Visualization of cross attentions.}
    We visualize the cross-attention of a single query, to all patches of all previous frames, for two intermediate blocks of the predictor.
    The selected blocks 6 and 16, zero-indexed, are those where the attention maps are the most informative.
    The model appears to track previous locations of the object of interest and its movement through the frames.
    Top row: five past frames in color and full resolution, and the future frame to be predicted low resolution and grayscale colors.
    The location of the query used for the visualization is marked in red in the future frame.
    Second and third rows: attention maps of the two selected blocks.
    The spatial location of the query is also marked in the attention maps for ease of visualization.
    }
    \label{fig:appendix-attn-visualization}
\end{figure}

\clearpage
In \Cref{fig:appendix-attn-visualization}, we visualize the cross-attention maps of a single query as it attends to all patches of all previous frames to predict the content of a future patch.
For the visualization, we select the two intermediate blocks where the attention maps are the most informative.
The model appears to track previous locations of the object of interest and its movement through the frames.

\clearpage
\subsection{Intuitive physics}
\label{appendix:intphys-all-results}

For completeness, in \Cref{tab:intphys-all-results}, we report the average accuracy per category for the IntPhys \citep{garrido2025intuitive}, GRASP \citep{jassim2024grasp}, and InfLevel \citep{weihs2022benchmarking} benchmarks.
These scores, averaged across categories, correspond to \Cref{tab:intphys-compact-results} in the main text.

\begin{table}[H]
    \centering
    \caption{
        \textbf{Intuitive physics evaluation.}
        Average accuracy per benchmark and category.
    }
    \label{tab:intphys-all-results}
    \begin{subtable}[t]{1\textwidth}
        \centering
        \small
        \caption{IntPhys benchmark, three categories.}
        \label{tab:intphys-by-category}
        \begin{tabular}{@{}lr ccc@{}}
            \toprule
                    & Encoder & Object Permanence & Unchangeableness & Spatiotemporal Continuity \\
            \midrule
            DINO-Foresight & ViT-B  & 78.3 & \phantom{0}90.0 & \phantom{0}95.0 \\
            V-JEPA  & ViT-L         & 85.0 & \phantom{0}98.3 & \phantom{0}93.3 \\  %
            V-JEPA  & ViT-H         & 78.3 & \phantom{0}95.0 & \phantom{0}95.0 \\  %
            \midrule
            COSMOS & 4B & 98.4 & 100.0 & 100.0 \\  %
            COSMOS & 12B & 98.4 & 100.0 & 100.0 \\  %
            \midrule
            DINO-world    & ViT-B         & 83.3       & \phantom{0}95.3 & \phantom{0}95.3 \\
            \bottomrule
        \end{tabular}
    \end{subtable}
    \begin{subtable}[t]{1\textwidth}
        \centering
        \tiny
        \caption{GRASP benchmark, sixteen categories, split in two rows for readability.}
        \label{tab:grasp-by-category}
        \begin{tabular}{@{}lr ccccccccc@{}}
            \toprule
            & Encoder & Collision & Continuity & Gravity & Gravity & Gravity & Gravity & Gravity & Inertia & Inertia2 \\
            &  &  &  &  & Continuity & Inertia & Inertia2 & Support &  &  \\
            \midrule
            DINO-Foresight & ViT-B  & \phantom{0}0.0 & 71.1 & 73.4 & 46.1 & 14.8 & 53.1 & 100.0 & 44.5 & 60.2 \\
            V-JEPA & ViT-L  & 54.7 & 47.7 & 55.5 & 64.1 & 75.0 & 53.1 & 100.0 & 47.7 & 53.9 \\
            V-JEPA & ViT-H  & 28.9 & 53.9 & 58.6 & 71.9 & 78.9 & 56.2 & 100.0 & 52.3 & 68.0 \\
            \midrule
            COSMOS & 4B  & 56.3 & 51.6 & 56.3 & 57.0 & 89.8 & 51.6 & 100.0 & 48.4 & 46.1 \\
            \midrule
            DINO-world & ViT-B  & \phantom{0}0.0 & 79.7 & 89.1 & 28.9 & 59.4 & 59.4 & 100.0 & 57.8 & 71.1 \\
            \bottomrule
        \end{tabular}\vspace{0.5em}
        \begin{tabular}{@{}ccccccc@{}}
            \toprule
            Object & Object & Object & Solidity & Solidity & Unchangeableness & Unchangeableness2 \\
            Permanence & Permanence2 & Permanence3 & Continuity & Continuity2 &  &  \\
            \midrule
            100.0 & 46.9 & 99.2 & 71.9 & 83.6 & 100.0 & 73.4 \\
            \phantom{0}27.3 & 82.0 & 99.2 & 87.5 & 60.9 & \phantom{0}91.4 & 71.9 \\
            \phantom{0}75.8 & 94.5 & 99.2 & 88.3 & 81.2 & \phantom{0}93.0 & 67.2 \\
            \midrule
            \phantom{00}0.0 & 53.9 & 96.9 & 79.7 & 23.4 & \phantom{0}88.3 & 61.7 \\
            \midrule
            100.0 & 83.6 & 99.2 & 96.9 & 93.0 & 100.0 & 98.4 \\
            \bottomrule
        \end{tabular}
    \end{subtable}
    \begin{subtable}[t]{1\textwidth}
        \centering
        \caption{InfLevel benchmark, three categories.}
        \label{tab:inflevel-by-category}
        \begin{tabular}{@{}lr ccc@{}}
            \toprule
                    & Encoder & Continuity & Gravity & Solidity \\
            \midrule
            DINO-Foresight & ViT-B  & 86.3 & 51.0 & 50.7 \\
            V-JEPA  & ViT-L         & 71.4 & 52.8 & 52.4 \\  %
            V-JEPA  & ViT-H         & 76.7 & 51.4 & 51.6 \\  %
            \midrule
            COSMOS & 4B             & 35.6 & 49.7 & 49.2 \\
            \midrule
            DINO-world    & ViT-B   & 88.5 & 51.1 & 51.3 \\
            \bottomrule
        \end{tabular}
    \end{subtable}
\end{table}

\end{document}